%% file: cav21_manthan.tex
\documentclass[runningheads]{llncs}

\usepackage{xcolor}
\usepackage[hidelinks,colorlinks = true,
           linkcolor = blue,
            urlcolor  = blue,
            citecolor = blue,
            anchorcolor = blue,
            bookmarks=false]{hyperref}
\usepackage{graphicx,subcaption}
\usepackage{enumitem}
\usepackage{amsmath}
\usepackage[labelfont=bf]{caption}
\usepackage{multicol}
\usepackage{multirow}
\usepackage{array}
\usepackage{booktabs}
\usepackage{tikz}
\usetikzlibrary{bayesnet}
\usetikzlibrary{arrows}
\usetikzlibrary{backgrounds,automata}
\usetikzlibrary{decorations.pathreplacing}
\usetikzlibrary{calc}
\tikzstyle{graphnode}=[circle,draw, fill=white, minimum size=2em]
\tikzstyle{decision} = [diamond, draw, text badly centered, inner sep=3pt]
\usepackage{subcaption}
\usepackage{tcolorbox}
\usepackage[noend]{algpseudocode}
\usepackage{algorithm}
\usepackage{tabularx}

\input{macros}

\usepackage[draft]{todonotes}   %

\begin{document}
\title{Engineering an Efficient Boolean Functional Synthesis Engine\thanks{{\tool} is available at \url{https://github.com/meelgroup/manthan}}}

\author{Priyanka Golia\inst{1,2} \and  Friedrich Slivovsky\inst{3} \and Subhajit Roy\inst{1} \and Kuldeep S. Meel\inst{2}}
\authorrunning{Golia, Slivovsky, Roy, and Meel}
\institute{ Indian Institute of Technology Kanpur  \and  National University of Singapore \and TU Wien
	\\
}

\maketitle              %
\begin{abstract}

	Given a Boolean specification between a set of inputs and outputs, the problem of Boolean functional synthesis is to synthesise each output as a function of inputs such that the specification is met. Although the past few years have witnessed intense algorithmic development, accomplishing scalability remains the holy grail.
	The state-of-the-art approach combines machine learning and automated reasoning to efficiently synthesise Boolean functions. In this paper, we propose four algorithmic improvements for a data-driven framework for functional synthesis: using a dependency-driven multi-classifier to learn candidate function, extracting uniquely defined functions by interpolation, variables retention, and  using lexicographic {\maxsat} to repair candidates.

	We implement these improvements in the state-of-the-art framework, called {\manthan}. The proposed framework is 
	called {\tool}. {\tool} shows significantly improved runtime performance compared to {\manthan}. In an extensive 
	experimental evaluation on 609 benchmarks, {\tool} is able to synthesise a Boolean function vector for 509 
	instances compared to 356 instances solved by {\manthan} -- an increment of 153 instances over the state-of-the-art. To put this into perspective, {\manthan} improved on the prior state-of-the-art by only 76 instances.
	
\end{abstract}

\input{chapters/introduction}
\input{chapters/preliminaries}

\input{chapters/approach}

\input{chapters/algorithm}

\input{chapters/experimental_results}

\input{chapters/relatedwork}
\input{chapters/conclusion}

\clearpage
\bibliographystyle{splncs04}
\bibliography{ref}
\clearpage
\input{chapters/appendix}

\end{document}

%% file: macros.tex
\newcommand{\tool}{\ensuremath{\mathsf{Manthan2}}}
\newcommand{\unique}{UNIQUE}

\newcommand{\manthan}{\ensuremath{\mathsf{Manthan}}}

\newcommand{\bfss}{BFSS}
\newcommand{\ctosyn}{C2Syn}
\newcommand{\cadet}{CADET}

\newcommand{\unsat}{{UNSAT}}
\newcommand{\unsatcore}{{UnsatCore}}
\newcommand{\sat}{{SAT}}
\newcommand{\maxsat}{{MaxSAT}}
\newcommand{\lexmaxsat}{{LexMaxSAT}}

\newcommand{\totalorder}{\textit{TotalOrder}}

\newcommand{\candidateskf}{\ensuremath{\mathsf{CandidateSkF}}}

\newcommand{\unidef}{\ensuremath{\mathsf{UniDef}}}

\newcommand{\findorder}{\ensuremath{\mathsf{FindOrder}}}
\newcommand{\checksat}{\ensuremath{\mathsf{CheckSat}}}

\newcommand{\getsamples}{\ensuremath{\mathsf{GetSamples}}}
\newcommand{\findunates}{\ensuremath{\mathsf{FindUnates}}}
\newcommand{\uniquecall}{\ensuremath{\mathsf{FindUniqueDef}}}
\newcommand{\univar}{univar}
\newcommand{\clustery}{\ensuremath{\mathsf{ClusterY}}}
\newcommand{\addedge}{\ensuremath{\mathsf{AddEdge}}}
\newcommand{\removenode}{\ensuremath{\mathsf{RemoveNode}}}
\newcommand{\khopneighbor}{\ensuremath{\mathsf{kHopNeighbor}}}
\newcommand{\chunk}{\ensuremath{\mathsf{chunk}}}		
\newcommand{\subsety}{\ensuremath{\mathsf{subsetY}}}			
\newcommand{\dependson}{\ensuremath{\mathsf{dependson}}}		
\newcommand{\createdt}{\ensuremath{\mathsf{CreateDecisionTree}}}

\newcommand{\Path}{\ensuremath{\mathsf{Path}}}

\newcommand{\leafnodes}{\ensuremath{\mathsf{LeafNodes}}}
\newcommand{\Label}{\ensuremath{\mathsf{Label}}}

\newcommand{\repairskf}{\ensuremath{\mathsf{RepairSkF}}}
\newcommand{\findrepaircandidate}{\ensuremath{\mathsf{FindRepairCandidates}}}

\newcommand{\findcore}{\ensuremath{\mathsf{FindCore}}}
\newcommand{\maxsatlist}{\ensuremath{\mathsf{MaxSATList}}}

%% file: chapters/introduction.tex
\section{Introduction}
\label{sec:introduction}
Given two sets $X = \{x_1,\ldots,x_n\}$ and  $Y = \{y_1,\ldots,y_m\}$ of variables and a Boolean formula $F(X,Y)$ over $X \cup Y$, the problem of Boolean functional synthesis is to compute a vector $\Psi = \langle \psi_1,\ldots,\psi_m \rangle$ of Boolean functions $\psi_i$ (often called \emph{Skolem functions}) such that $\exists Y F(X,Y) \equiv F(X, \Psi(X))$. Informally, given a specification
between inputs and outputs, the task is to synthesise a function $\Psi$ that maps each assignment of the inputs to an assignment of the outputs so that the combined assignment meets the specification (whenever such an assignment exists).
With origins tracing to Boole's seminal work~\cite{boole1847}, functional synthesis is a fundamental problem in computer science that has a wide variety of applications in areas such as  circuit synthesis~\cite{KS00}, program synthesis~\cite{SGF13}, automated program repair~\cite{JMF14}, cryptography~\cite{MM20}, logic minimization~\cite{B89,BS89}. For example, the relation $F$ can specify the allowed behavior of a circuit of interest and the function $\Psi$ corresponds to the implementation of the desired circuit. As pointed out by Jiang, Lin, and Hung~\cite{JLH09}, relations can succinctly capture the conventional notion of {\em don't cares}. Furthermore, extracting functions from Boolean relations also has applications in 2-level logic minimization under the Sum-of-Products (SOP) representation~\cite{GDN92,JS92,LS90}.

Over the past two decades, functional synthesis has seen a surge of interest, leading to the development of new approaches that can be broadly classified into three categories:
1) incremental determinization iteratively identifies variables with unique Skolem functions and takes ``decisions" on any remaining variables by adding temporary clauses that make them deterministic~\cite{R19,RS16,RTRS18}.
2) Skolem functions can be obtained by eliminating quantifiers using functional composition, and Craig interpolation can be applied to reduce the size of composite functions~\cite{JLH09,J09}.
Although this typically does not scale to large specifications, it was shown to work well using ROBDDs in combination with carefully chosen variable orderings~\cite{FTV16,TV17}.
3) CEGAR-style approaches start from an initial set of approximate Skolem functions, followed by a phase of counter-example guided refinement to patch these candidate functions~\cite{JSCTA15,ACJS17,ACGKS18}.
With the right choice of initial functions, the CEGAR phase can often be skipped entirely, a phenomenon that can be analyzed in terms of knowledge compilation~\cite{ACGKS18,AACKRS19}.

Recently, we proposed a new data-driven approach {\manthan} \cite{GRM20}. {\manthan} relies on constrained sampling~\cite{SGM20} to generate satisfying assignments of the formula~$F$, which are fed to a decision-tree learning technique such that the learned classifiers represent potential Skolem functions, called candidates.
The candidates are repeatedly tested for correctness and repaired in a subsequent CEGAR loop, with a MaxSAT solver minimizing the number of repairs required for each counterexample. While {\manthan} achieved significant improvement of the state-of-the-art, a large number of problems remain beyond its reach (and other synthesis engines).

The primary contribution of this work is to address scalability barriers faced by {\manthan}. To this end, we propose following crucial algorithmic innovations:

\begin{enumerate}
	\item \textbf{Interpolation-based Unique Function Extraction}: We identify a subset of variables with {\em unique Skolem function} and extract these functions by interpolation, thereby reducing the number of functions that need to be learned.
	
	\item \textbf{Clustering-based Multi-Classification}: We propose a clustering-based approach that can take advantage of multi-classification to learn candidate functions for sets of variables at a time. 
	
	\item \textbf{Learning and Repair over Determined Features}: 
	 Whenever it is determined that a candidate function for a variable is indeed a Skolem function, we do {\underline{not}} substitute for and eliminate this variable, and instead retain it as a possible feature during learning and repair. Our strategy stands in stark contrast to the conventional wisdom that advocates variable elimination.

      \item \textbf{Lexicographic {\maxsat}-based Dependency-Aware Repair}: We design a lexicographic {\maxsat}-based strategy for identifying repair candidates so as to take into account dependencies among candidate functions.

\end{enumerate}

To measure the impact of these proposed algorithmic innovations, we implemented them in a system named {\tool} and performed an extensive evaluation on a benchmark suite used in prior studies~\cite{AACKRS19,ACGKS18,GRM20}.
In terms of solved instances, the results are decisive.
Out of 609 instances, {\manthan} and CADET are able to solve 356 and 280 instances, in line with experimental results reported in prior work that saw a 76 instance lead of {\manthan} over the then state-of-the-art CADET~\cite{GRM20}.
{\tool} solves 509 instances and thereby achieves a dramatic improvement of 153 instances over {\manthan}, more than doubling the substantial increase in the number of solved instances achieved by {\manthan} over {\cadet}.

The rest of the paper is organized as follows: In Section~\ref{sec:notation-preliminaries}, we first introduce notation and then provide some background on {\manthan}.In Section~\ref{sec:overview}, we present an overview of the invocations implemented in {\tool}, before giving a detailed algorithmic description in Section~\ref{sec:algorithm}. We then describe the experimental methodology and discuss results with respect to each of the technical contributions of {\tool} in Section~\ref{sec:experimental-evaluations}. We cover related work in Section~\ref{sec:relatedwork}. Finally, we conclude in Section~\ref{sec:conclusion}.

%% file: chapters/preliminaries.tex
\section{Background}
\label{sec:notation-preliminaries}
We use lower case letters to denote propositional variables and capital letters to denote sets of variables.
Given a set $\{v_1, \dots, v_n\}$ of variables and $1 \leq i \leq j \leq n$, we write $V_{i}^{j}$ for the subset $\{v_i, v_{i+1}, \ldots, v_{j}\}$.
We use standard notation for logical connectives such as $\land, \lor$ and $\lnot$.
A \emph{literal} is a variable or a negated variable.
A formula $\varphi$ is in Conjunctive Normal Form (CNF) if it is a conjunction of \emph{clauses}, where each \emph{clause} is a disjunction of literals.
We write $Vars(\varphi)$ to denote the set of variables used in $\varphi$. 
A \emph{satisfying assignment} of a formula $\varphi$ is a mapping $\sigma: Vars(\varphi) \rightarrow \{0,1\}$ such that $\varphi$ evaluates to True under~$\sigma$.
We write $\sigma \models \varphi$ to denote that $\sigma$ is a satisfying assignment of $\varphi$.
Given a subset $V$ of variables, we write $\sigma[V]$ to denote the restriction of $\sigma$ to $V$.
An \emph{unsatisfiable core} of a formula in CNF is a subset of clauses for which there is no satisfying assignment. We use {\unsatcore} to denote an unsatisfiable core when the formula is understood from the context.

For a given CNF formula in which some clauses are declared as \emph{hard constraints} and the rest are declared as \emph{soft constraints}, the problem of (partial) {\maxsat} is to find an assignment of the given formula that satisfies all hard constraints and maximizes the number of satisfied soft constraints.
Furthermore, \emph{lexicographic} partial {\maxsat}, or {\lexmaxsat} for short, is a special case of partial {\maxsat}  in which there is a preference in the order in which to satisfy the soft constraints. 

\subsection{Functional Synthesis}
We assume a relational specification $\exists Y F(X,Y)$ such that $X = \{x_1,\ldots, x_n\}$ and $Y = \{ y_1, \ldots, y_m\}$.
We write $F$ and $F(X,Y)$ interchangeably, and use $F(X,Y)|_{y_i=b}$ to denote the result of substituting $b \in \{0, 1\}$ for $y_i$ in $F(X,Y)$.

\paragraph{\textbf{Problem Statement:}} Given a specification $\exists Y F(X,Y)$ with inputs~$X$ and outputs $Y$, the task of \emph{Boolean functional synthesis} is to find a function vector $\Psi = \langle \psi_1, \ldots, \psi_m \rangle$ such that $\exists F(X,Y) \equiv F(X,\Psi(X))$.
We refer to $\Psi$ as a \emph{Skolem function vector} and to the function $\psi_i$ as a \emph{Skolem function} for $y_i$.

We solve a slightly relaxed version of this problem by synthesising a Skolem function vector $\langle \psi_1,\ldots,\psi_m\rangle$ such that $y_i$ = $\psi_i(X,y_1,\ldots,y_{i-1})$ for a given order $y_1,\ldots,y_m$; this is ultimately equivalent to synthesising $\Psi(X)$, since each $\psi_i(X,y_1, \ldots, y_{i-1})$ can be transformed into a function depending only on~$X$ by substituting the functions for $y_1,\ldots,y_{i-1}$. We write $\prec_{d}$ to denote the (smallest) partial order on the output variables $Y$ such that $y_i  \prec_{d} y_j$ if $y_j$ appears in $\psi_i$, and say that \emph{$y_i$ depends on $y_j$} whenever $y_i  \prec_{d} y_j$.

\subsection{Definability}

\begin{definition}[\cite{LM08}]
	Let $F(W)$ be a formula, $w \in W$, $S \subseteq W \setminus w$. $F(W)$ \emph{defines} $w$ in terms of $S$ if and only if there exists a formula $H(S)$ such that $F(W) \models w \leftrightarrow H(S)$. In such a case, $H(S)$ is called a \emph{definition} of $w$ on $S$ in $F(W)$.
\end{definition}

To this end, given $F(W)$ defined on $W = \{w_1, w_2, \ldots w_n\}$. We create another set of {\em fresh} variables $Z = \{z_1, z_2, \ldots z_n\}$. Let $F(W \mapsto Z)$ represent the formula where every $w_i \in W$ in $F$ is replaced by $z_i \in Z$.

\begin{lemma}[Padoa's Theorem]\label{lemma:padoa}
	\begin{align*}	
	\text{Let, } I(W,Z,S,i) = F(W) \wedge F(W \mapsto Z) \wedge \left(\bigwedge_{w_j \in \mathcal{S}; j\neq i} (w_j \leftrightarrow z_j) \right) \\ \wedge w_i \wedge \neg z_i
	\end{align*} 
	$F$ defines $w_i \in W$  in terms of $S$  if and only if $I(W,Z,S,i)$ is {\unsat}. 
\end{lemma}

\subsection{Manthan: Background}
We now give a brief overview of the state-of-the-art Boolean functional synthesis tool {\manthan}~\cite{GRM20}.
Given a specification, {\manthan} computes Skolem functions in several phases described below.
	
\paragraph{\bfseries Preprocessing:}  A variable $y_i$ is \emph{positive unate} (resp. \emph{negative unate}) in $F(X,Y)$, if $F(X,Y)|_{y_i=0} \land \lnot F(X,Y)|_{y_i=1}$ (resp. $F(X,Y)|_{y_i=1} \land \lnot F(X,Y)|_{y_i=0}$) is {\unsat}~\cite{ACGKS18}.
The Skolem function for a positive unate (resp. negative unate) variable $y_i$ is the constant function $\psi_i = 1$ (resp. $\psi_i = 0$). {\manthan} finds unates as a preprocessing step.
	
\paragraph{\bfseries Learning Candidates:} {\manthan} adopts an \emph{adaptive} weighted sampling strategy to sample satisfying assignments of $F(X,Y)$, which are used to learn decision tree classifiers.
More specifically, {\manthan} samples uniformly over the input variables $X$ while biasing the output variables $Y$ towards a particular value. With the data generated, {\manthan} learns approximate candidate functions using a dependency driven binary classifier. To learn a candidate function $\psi_i$ corresponding to $y_i$, {\manthan} considers the value of $y_i$ in a satisfying assignment as a label, and values of $X \cup \hat{Y}$ as a feature set to construct a decision tree $dt$, where $\hat{Y}$ is the set of $Y$ variables, such that for $y_j$ of $\hat{Y}$, $y_j \not \prec_{d} y_i$.
From the learned decision tree $dt$, {\manthan} obtains the candidate function as the disjunction of all the paths with leaf node label $1$. For every $y_k$ occurring as decision node in $dt$, {\manthan} updates the dependencies as $y_i \prec_{d} y_k$. Finally, {\manthan} extends the partial order $\prec_{d}$ to get a {\totalorder} of $Y$ variables.
	
\paragraph{\bfseries Verification:}
{\manthan} checks if the learned candidates are Skolem functions or not by checking satisfiability of the \emph{error formula} $E(X,Y,Y')$ defined as
	\begin{align}
	\label{eqn:errorformula}
	E(X,Y,Y') = F(X,Y) \land \lnot F(X,Y') \land (Y' \leftrightarrow \Psi),
	\end{align}
where $Y' = \{y_1', \dots, y_m'\}$ is a set of fresh variables. It is readily verified that $\Psi$ is a Skolem function vector if, and only if, $E(X,Y,Y')$ is {\unsat}~\cite{JSCTA15}.	If $E(X,Y,Y')$ is {\sat} and $\sigma \models E(X,Y,Y')$, then {\manthan} has a counterexample {$\sigma$} to fix. 
	
\paragraph{\bfseries Repairing Candidates:} {\manthan} finds candidate functions to repair by making a {\maxsat} call with hard constraints $F(X,Y) \land (X \leftrightarrow \sigma[X])$ and soft constraints $(Y' \leftrightarrow \sigma[Y'])$.
The output variables associated with the soft constraints that are \emph{not} satisfied form a smallest subset of output variables whose candidate functions need to change to satisfy the specification.
Now, to repair a candidate function $\psi_i$ corresponding to output variable $y_i$, {\manthan} constructs another formula $G_i(X,Y)$ as
	\begin{align}
	\label{eqn:Gformula}
	G_{i}(X,Y) := (y_i \leftrightarrow \sigma[y'_{i}]) \land F(X,Y) \land (X \leftrightarrow \sigma[X]) \land (\hat{Y} \leftrightarrow \sigma[\hat{Y}]),
	\end{align}
where $\hat{Y} \subset Y$ is the set $\hat{Y} =  \{ \totalorder[index(y_i)+1],\cdots,\totalorder[|Y|]\}$.
        
If $G_i(X,Y)$ turns out to be {\unsat}, then {\manthan} constructs a repair formula $\beta$ as the conjunction of all unit clauses of an {\unsatcore} of $G_i(X,Y)$.
Depending on the current valuation of the candidate function $\psi$, {\manthan} strengthens or weaken the candidate by the repair formula $\beta$.
Otherwise, if $G_i(X, Y)$ is SAT, {\manthan} looks for other candidate functions to repair instead.
	
During the repair phase, {\manthan} uses self-substitution~\cite{J09} as a fallback: whenever more than 10 iterations are needed for repairing a particular candidate, {\manthan} directly synthesises a Skolem function for that variable via self-substitution.

%% file: chapters/approach.tex
\section{Overview}
\label{sec:overview}

In this section, we provide an overview of our primary contributions in {\tool}, building on the {\manthan}~\cite{GRM20} infrastructure.

\subsection{Interpolation-based Unique Function Extraction}
In order to reduce {\manthan}'s reliance on data-driven learning, we seek to identify a subset $Z \subseteq Y$ and the corresponding Skolem function vector $\Phi$ such that $\Phi$ can be extended to a valid Skolem function vector $\Psi$. In the following, we call such a $Z$ a {\em determined set}. Observe that unate variables form a determined set $Z$.
To grow $Z$ further, we rely on the notion of definability, and iteratively identify the variables $y_i \in Y$ such that $y_i$ is definable in terms of rest of the variables such that its definition $\psi_i$ respects the dependency constraints imposed by the definitions of variables in $Z$. To extract the corresponding definitions, we rely on the Padoa's theorem (Lemma~\ref{lemma:padoa}) to check whether $y_i$ is definable in terms of rest of the variables and then employ interpolation-based extraction of the corresponding definition~\cite{F20}. %

The usage of unique function extraction significantly reduces the number of variables for which {\tool} needs to learn and repair the candidates since unique functions do not need to undergo refinement. 
While our primary motivation for unique function extraction was to reduce the over-reliance on learning, it is worth emphasizing that interpolation-based extraction could also compute functions with large size; these functions would require a prohibitive number of samples and as such lie beyond the scope of a practical learning-based technique.

We close by highlighting the importance of allowing $y_i$ to depend, subject to dependency constraints, on other $Y$ variables. Consider, $X= \{x_1\}$ and $Y = \{y_1,y_2\}$, and let $F(X,Y) := (y_1 \lor y_2 ) \land (\lnot y_1 \lor \lnot y_2)$. Neither $y_1$ nor $y_2$ is defined by $x_1$. But $y_2$ is  definable in terms of $\{y_1\}$ (and therefore, also $\{x_1, y_1\}$) with its corresponding Skolem function $\psi_2(x_1,y_1):= \lnot y_1$.

{\bfseries Impact:} For over $40\%$ of our benchmarks, {\tool} could extract Skolem functions for at least $95\%$ of total variables via unique function extraction.

\subsection{Learning and Repair over Determined Features}
As mentioned in the previous section, we focus on constructing a determined set $Z$ consisting of unates and variables with unique functions. All the variables in $Z$ can be eliminated by substituting them with their corresponding definitions (in case of unates, the definition is a constant: True or False).
Variable elimination has a long history as an effective preprocessing strategy~\cite{AACKRS19,ACGKS18,BLS11,GRM20}, and, following this tradition, {\manthan} performs variable elimination wherever possible.
In particular, it eliminates unates as well as variables for which definitions can be obtained via syntactic gate extraction techniques.

While substituting for variables in $Z$ does not affect the \emph{existence} of Skolem functions for variables $y_i \in Y \setminus Z$, the \emph{size} of these functions can increase substantially when they are not allowed to depend on variables in $Z$.
We also observe that variables in $Z$ can considered as \emph{determined features} and the Skolem functions for some $y_i \in Y \setminus Z$ can be efficiently represented in $Z$.
For example, consider the following scenario: let $X = \{x_1, x_2\}$, $Y = \{y_1,y_2\}$ and $F(X,Y) = (y_1 \lor y_2) \land (\lnot y_1 \lor \lnot y_2) \land (y_1 \leftrightarrow (x_1 \oplus x_2))$. Observe that the Skolem function for $y_2$  in terms of $X$ in the transformed formula $F(x_1,x_2,y_2)$ will have to be learned as $ \lnot (x_1 \oplus x_2)$. However, when allowing learning over $y_1$, then the desired Skolem function for $y_2$ can simply be learned as $\lnot y_1$.\footnote{There is an analogy with the role of \emph{latent features} in machine learning, which allow for the compact representation of a model but must first be computed from observable features: elimination of variables with unique Skolem functions turns observable features into latent features that must be recovered by the learning algorithm.}

Further, every iteration of our repair phase adds clauses over the literals in the formula, and therefore allowing a repair clause to contain a variable $y_i \in Z$ with definition $\psi_i$ increases the expressiveness of the clauses during the repair phase, akin to bounded variable addition.

We conclude that, contrary to conventional wisdom, variables in the determined set $Z$ should not be eliminated and instead should be retained as features for the learning and repair phases of {\tool}.

{\bfseries Impact}: The retention of variables in the determined set allows {\tool} to solve $25$ more benchmarks.

\subsection{Clustering-based Multi-Classification}
For some of the benchmarks, {\manthan} spends~$\sim 74\%$ of its time in learning the candidate functions. To reduce this learning time, {\tool} uses the following strategy:
\begin{enumerate}
	\item Partition the set of $Y$ variables into disjoint subsets,
	\item Use a multi-classifier (instead of a binary-classifier) to learn candidate Skolem functions for each partition. 
\end{enumerate}

\begin{figure}
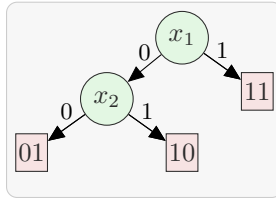


	{
		\centering
	\scalebox{1}{\tikz{
			\node[obs,fill=green!10,minimum size=0.7cm] (x1) {$x_1$};%
			\node[latent,below=of x1,yshift=0.9cm, xshift=-1cm,fill=green!10,minimum size=0.7cm] (x2) {$x_2$}; %
			\node[latent, rectangle,below=of x1, yshift=0.9cm, xshift= 1cm,fill=red!10, minimum width=0.2cm,
			minimum height = 0.5cm] (l1) {$11$}; %
			\node[latent, rectangle, below=of x2, minimum size=0.7cm, yshift=0.9cm, minimum width=0.2cm,
			minimum height = 0.5cm,xshift= -1cm,fill=red!10] (l2) {$01$}; %
			\node[latent, rectangle, below=of x2, minimum size=0.7cm, yshift=0.9cm, minimum width=0.2cm,
			minimum height = 0.5cm, xshift= 1cm,fill=red!10] (l3) {$10$}; %
			\plate [fill=gray!30,opacity=0.2] {plate1} {(x1)(x2)(l1) (l2) (l3)} {}; %
			\edge {x1}   {l1}
			\edge {x2}   {l2}
			\edge {x2}   {l3}
			
			\path[->,draw]
			(x1) edge node [above] {$0$} (x2)
			(x1) edge node[above] {$1$} (l1)
			(x2) edge node[above] {$0$} (l2)
			(x2) edge node[above] {$1$} (l3)
			
		}}
		\caption{Multi-Classification: learned decision tree with labels $\{y_1,y_2\}$ and features $\{x_1,x_2\}$}\label{fig:example-multiclass}
		
	}
\end{figure}

For example, let $X = \{x_1,x_2\}$ and $Y = \{y_1,y_2\}$ in $\exists Y F(X,Y)$. Figure~\ref{fig:example-multiclass} shows the learned decision tree with labels $\{y_1,y_2\}$, and features $\{x_1,x_2\}$. The expected number of classes to learn two $Y$ variables is $2^2=4$, but as shown in Figure~\ref{fig:example-multiclass}, the decision tree classifies the labels into $3$ classes $\langle 01,10,11 \rangle$. The candidate function $\psi_1$ corresponding to $y_1$ is the disjunction of paths from root to leaf node with label of $y_1$ being 1, i.e, the classes $10$ and $11$. Hence, the candidate function $\psi_1:=(\lnot x_1 \land x_2) \lor (x_1)$. Similarly, the candidate function $\psi_2$ for $y_2$ is $\psi_2:= (\lnot x_1 \land \lnot x_2) \lor (x_1)$.

The candidate Skolem function for a variable $y_i$ of a chosen subset is obtained as the disjunction of all the paths from the root to leaf node with a label of $y_i$ being $1$. We further update the partial dependency as $y_i \prec_{d} y_j$,  for all $y_j$ variables occurring in $\psi_i$. Now, let us consider the case with two different subsets $\{y_1,y_2\}$ and $\{y_3, y_4\}$, and also assume that $y_1 \prec_{d} y_3$, then the feature set to learn $\{y_3,y_4\}$ would be $\{X, y_2\}$. The feature set to learn a chosen subset would include a variable $y_j$, only if $y_j \not \prec_{d} y_i$ for every variable $y_i$ of the subset.

An important question that remains to be answered is \emph{how should the variable partitioning be driven}? The intuition behind our approach lies in the fact that low cohesion among variables in a partition would impose fewer constraints, leading to larger trees and multiplying the number of classes. Therefore, in some sense, we would like to learn \emph{related} variables together. {\tool} uses the distance in the \emph{primal graph}~\cite{SS10} to cluster $Y$ variables into disjoint subsets, such that variables in a subset are closely related.

{\bfseries Impact}: We observe a decrease of 252 seconds in the PAR-2 score by using a multi-classifier to learn a subset of variables together over learning one candidate at a time.

\subsection{Lexicographic {\maxsat}-based Dependency-Aware Repair}
\label{sec:lexo-maxsat}

Let us start by demonstrating a troublesome scenario for {\manthan} on the same running example as above: $X = \{x_1, x_2\}$ and $Y = \{y_1,y_2\}$ and let $F(X,Y) = (y_1 \lor y_2) \land (\lnot y_1 \lor \lnot y_2)$ in $\exists Y F(X,Y)$, with the candidates $\psi_1 = 1$ and $\psi_2 = 1$, and {\totalorder} = $\{y_1, y_2 \}$. As the candidates are not yet Skolem functions, {\manthan} starts off by identifying a candidate for repair by invoking {\maxsat} with hard constraints $F(X,Y) \land (X \leftrightarrow \sigma[X])$ and soft constraints $(y_1 \leftrightarrow 1) \land (y_2 \leftrightarrow 1)$, where $\sigma$ is a satisfying assignment of the error formula~(\ref{eqn:errorformula}).
As either $y_1$ or $y_2$ can be flipped to fix the counterexample $\sigma$, let us assume {\maxsat} does not satisfy the soft constraint $(y_2 \leftrightarrow 1)$, thereby selecting $\psi_2$ for repair.

In order to repair $\psi_2$, {\manthan} constructs the formula $G_2$~(\ref{eqn:Gformula}) as $G_2 = F(X,Y) \land (X \leftrightarrow \sigma[X]) \land (y_2 \leftrightarrow 1)$. As $G_2$ is not allowed to constrain over $y_1$, it turns out as {\sat}, hence adding $\psi_1$ as a candidate to repair. Therefore, to fix the counterexample $\sigma$, {\manthan} fails to repair the candidate, and requires an additional repair iteration. This scenario could have been averted and the counterexample $\sigma$ can be fixed in the same repair iteration if $\psi_1$ was selected before $\psi_2$

{\tool} uses {\lexmaxsat}~\cite{IMM18} to satisfy the soft constraints in accordance to the {\totalorder}. For the aforementioned problem, if the soft constraint $y_1 \leftrightarrow 1$ takes preference over $y_2 \leftrightarrow 1$, {\tool} would pick candidate corresponding to $y_1$ as a repair candidate. Therefore, the use of {\lexmaxsat} in finding repair candidates reduces the required number of iterations to fix a counterexample.

However, {\lexmaxsat} can be expensive~\cite{ABGL13,MAGL11}.
To avoid frequent {\lexmaxsat} calls, {\tool} first computes a list of candidates to repair using unweighted {\maxsat}.
This list can grow whenever a formula $G_i$ turns out to be {\sat}. Once its size exceeds a certain threshold, {\tool} recomputes another set of repair candidates using {\lexmaxsat}. In particular, {\lexmaxsat} is used only if {\tool} has to fix many candidates in a single repair iteration due to an ordering constraint.  

{\bfseries Impact}: We observe a decrease of more than $100$ seconds in the PAR-2 score by using {\lexmaxsat}.

%% file: chapters/algorithm.tex
\section{Algorithm}
\label{sec:algorithm}
In this section, we present a detailed algorithmic description of {\tool}. It takes a formula $F(X,Y)$, and returns a Skolem function vector. {\tool} considers fixed values for $k$ and $s$, where $k$ is the maximum edge distance that is used to cluster $Y$ variables together, and $s$ is the maximum number of $Y$ variables that can be learned together.

\input{algorithms/main_algo.tex}

{\tool} is presented in Algorithm~\ref{algo:main-algo}, it starts off by extracting Skolem functions for unates and uniquely defined variables of the formula $F(X,Y)$ at line~\ref{line:mainalgo:preprocess}. The set $U$ represents all the $Y$ variables that are either unate or have unique Skolem functions. 
At line~\ref{line:mainalgo:getsample}, {\tool} generates the required number of samples. Next, at line~\ref{line:mainalgo:cluster}, {\tool} calls subroutine {\clustery} to cluster the $Y$ variables that are not in $U$. {\clustery} returns a list, \textit{$\subsety$}, that represents different subsets of $Y$ variables for which the candidates would be learned together. To learn the candidate functions for each subsets, {\tool} calls subroutine {\candidateskf} at line~\ref{line:mainalgo:candidateskf}. {\candidateskf} also updates the dependencies among $Y$ variables as per the learned candidate functions. {\tool} now finds a total order {\totalorder} of $Y$ variables in accordance with dependencies among the $Y$ variables at line~\ref{line:mainalgo:totalorder}. {\tool} then checks the satisfiability of the error formula $E(X,Y,Y')$, and if $E(X,Y,Y')$ is {\sat}, it calls subroutine {\findrepaircandidate} to find the list of candidates to repair at  line~\ref{line:mainalgo:repaircandidates}. Then at line~\ref{line:mainalgo:repair}, it calls subroutine {\repairskf} to repair the candidates. This process is continued until the error formula $E(X,Y,Y')$ is {\unsat}, and then, {\tool} returns a Skolem function vector.
Note that if $U=Y$, that is, if all $Y$ variables are either unate or uniquely defined, then {\tool} terminates after {\unidef}.

{\tool} uses subroutines {\getsamples}, {\findorder} and {\repairskf} as described in~\cite{GRM20}.\footnote{Note that the subroutines {\findrepaircandidate}, and {\repairskf} are referred to as {\maxsatlist} and $\mathsf{RefineSkF}$ in~\cite{GRM20}.} And like {\manthan}, {\tool} uses self-substitution~\cite{J09} as a fallback (see Section~\ref{sec:notation-preliminaries}).
We will now discuss the newly introduced subroutines.

\subsubsection{{\unidef}} Algorithm~\ref{algo:preprocess} presents the subroutine {\unidef}.
It assumes access to the following two subroutines:
\begin{enumerate}
			\item {\findunates}, which takes a formula $F(X,Y)$ as input and returns a list of unates and their corresponding Skolem functions. 
			\item {\uniquecall}, which takes a formula $F(X,Y)$, a variable $y_i$, and a defining set $X, y_1, \ldots,y_{i-1}$ as input, and determines whether the given variable $y_i$ is defined with respect to the defining set or not. If the variable $y_i$ is defined, {\uniquecall} returns true, along with the extracted definition $\psi_i$. Otherwise, it returns false (and an empty definition). 
\end{enumerate}
	
	 {\unidef} first calls {\findunates} to find the unates and their corresponding Skolem functions at line~\ref{line:algo:preprocess}. Then, it calls subroutine {\uniquecall} with defining set $\{X,y_1,\ldots,y_{i-1}\}$ for each existentially quantified variable $y_i$ which is not unate at line~\ref{line:algo:preprocess:uniquecall}. If {\uniquecall} returns true, {\unidef} adds $y_i$ to the set \emph{\univar} at line~\ref{line:algo:preprocess:addunivar}. {\unidef} adds variables occurring in $\psi_i$ to the list ${\dependson[y_i]}$ at line~\ref{line:algo:preprocess:dependson}.

\input{algorithms/preprocess.tex}	
	\subsubsection{{\clustery}} Algorithm~\ref{algo:multiclass-clustering} presents the subroutine {\clustery}, it takes formula the $F(X,Y)$, $k:$ an edge distance parameter, $s:$ maximum allowed size of a cluster of $Y$ variables, and $U:$ list of unate and uniquely defined $Y$ variables, and it returns a list of all subsets of $Y$ that would be learned together. {\clustery} assumes access to a subroutine {\khopneighbor}, which takes a graph, a variable $y$, and an integer $k$ as input, and returns all variables within distance $k$ of $y$ in the graph.
	 {

           {\clustery} first creates a graph \emph{graph} with $Y \setminus U$ as vertex set and edges between variables $y_i$ and $y_j$ that share a clause in $F(X,Y)$. %
{\clustery} then calls subroutine {\khopneighbor} for each variable $y_i$. The set of variables returned by {\khopneighbor} is stored as {\chunk}. If the size of {\chunk} is greater than $s$, {\clustery} reduces the value of $k$ by one at line~\ref{line:algo:cluster:reducek}, and calls {\khopneighbor} again with the updated value of $k$. Otherwise, {\clustery} adds {\chunk} to {\subsety} at line~\ref{line:algo:cluster:addchunk}. Finally at line~\ref{line:algo:cluster:removenode}, {\clustery} removes the nodes corresponding to each variable of {\chunk} from \emph{graph}.
	 
	 \input{algorithms/cluster.tex}

	\subsubsection{{\candidateskf}} Algorithm~\ref{algo:multiclass-learning} presents the subroutine {\candidateskf}, it takes a set $\Sigma$ of samples, $F(X,Y)$, $\Psi$: a candidate function vector, {\chunk}: the set of variables to learn candidates, and {\dependson}: a partial dependency vector as input, and finds the candidates corresponding to each of the variables $y_i$ in {\chunk}. {\candidateskf} assumes access to subroutines {\createdt} and {\Path} as described by Golia et al.~\cite{GRM20}. The following are the additional subroutines used by {\candidateskf}.
		\begin{enumerate}	
			\item{\leafnodes}, which takes a decision tree $dt$ as an input and returns a list of leaf nodes of $dt$.
			\item{\Label($y_i,l$)}, which takes a variable $y_i$ and a leaf node $l$ as input, and returns $1$ if the class label corresponding to the node $l$ has value $1$ at the $i^{th}$ index.  	
		\end{enumerate}
		
	\input{algorithms/classification.tex}

{\candidateskf} starts off by initializing the set \emph{featset} of features with the set $X$ of input variables. It then attempts to find a list $D$ of variables $y_j$ such that $y_j \prec_{d} y_i$ where $y_i$ belongs to {\chunk}. Next, {\candidateskf} adds $Y \setminus D$ to \emph{featset}, and creates a decision tree $dt$ using samples from $\Sigma$ over \emph{featset} to learn the {\chunk} variables.  For a leaf node $l$ of $dt$, if {\Label($y_i,l$)} returns $1$, then $\psi_i$ is updated with the disjunction of the formula returned by subroutine {\Path}. Finally, {\candidateskf} iterates over all $y_j$ occurring in $\psi_i$ to add them to the list {\dependson[$y_i$]}.

	\subsubsection{{\findrepaircandidate}}Algorithm~\ref{algo:maxsat} presents the {\findrepaircandidate}. It starts with a {\lexmaxsat} call using hard constraints $F(X,Y) \land (X \leftrightarrow \sigma[X])$, soft constraints $(y_i \leftrightarrow \sigma[y'_i])$ for each $y_i$ of $Y$. The preference order on soft constraints is given by {\totalorder}. {\findrepaircandidate} calls the {\maxsatlist} subroutine, which returns a list of $Y$ variables \emph{ind} such that the soft constraints corresponding to variables in \emph{ind} were not satisfied by the optimal solution returned by the {\lexmaxsat} solver.
	
	\input{algorithms/maxsat.tex}

\input{chapters/example.tex}

%% file: algorithms/main_algo.tex
\begin{algorithm}[!b]
	
	\caption{\label{algo:main-algo}\tool(F(X,Y))}
	\begin{algorithmic}[1]
		
		\State $\Psi \gets \{ \psi_1 = \emptyset,\ldots, \psi_{|Y|} =\emptyset\} $ 
		
		\State {\dependson} $\gets$ $\{\}$
		
		\State U, $\Psi$, {\dependson} $\gets$ {\unidef}(F(X,Y),$\Psi$,{\dependson}) \label{line:mainalgo:preprocess}
		
		\State $\Sigma \gets$ {\getsamples}(F(X,Y))\label{line:mainalgo:getsample}
				
		\State  {\subsety} $\gets$ {\clustery}(F(X,Y),k,s,U) \label{line:mainalgo:cluster}
		
		\For {each {\chunk} $\in$ {\subsety}}
				\State $\Psi$, {\dependson} $\gets$ {\candidateskf}($\Sigma$,F(X,Y), $\Psi$, {\chunk}, {\dependson})\label{line:mainalgo:candidateskf}
		\EndFor
		
		\State {\totalorder} $\gets$ {\findorder}(\dependson)\label{line:mainalgo:totalorder}
		
		\Repeat 
			\State $E(X,Y,Y') \gets F(X,Y) \land  \lnot F(X,Y') \land (Y' \leftrightarrow \Psi)$
			\State ret, $\sigma \gets $ {\checksat}(E(X,Y,Y'))
			\If {ret = {\sat}} 
				\State ind $\gets$ {\findrepaircandidate}(F(X,Y),$\sigma$,{\totalorder})\label{line:mainalgo:repaircandidates}
				\For {$y_k \in $ ind }
					\State $\Psi  \gets$ {\repairskf}(F(X,Y),$\sigma$,$\Psi$,{\totalorder})\label{line:mainalgo:repair}
				\EndFor
			\EndIf
		\Until{ret = {\unsat}}
		\State \Return $\Psi$
	\end{algorithmic}
\end{algorithm}

%% file: algorithms/preprocess.tex
\begin{algorithm}[t]
	
	\caption{\label{algo:preprocess}{\unidef}(F(X,Y),$\Psi$,{\dependson})}
	\begin{algorithmic}[1]
	 \State	$\Psi$, unates $\gets$ {\findunates}(F(X,Y)) \label{line:algo:preprocess}
	 \State {\univar} $\gets$ $\emptyset$
	 \For {$y_i \in Y \setminus$ unates}
	 		\State definingvar $\gets X \cup \{ y_1, \ldots,y_{i-1}\}$
	 		\State ret, def $\gets$ {\uniquecall}(F(X,Y),$y_i$,definingvar) \label{line:algo:preprocess:uniquecall}
	 		\If {ret = true}
	 			\State {\univar} $\gets$ {\univar} $\cup$ $y_i$ \label{line:algo:preprocess:addunivar}
	 			\State $\psi_i \gets$ def \label{line:algo:preprocess:setdef}
	 			\For {$y_j \in \psi_i$}
	 				\State {\dependson[$y_i$]} $\gets$  {\dependson[$y_i$]} $\cup$ $y_j$ \label{line:algo:preprocess:dependson}
	 			\EndFor
	 		\EndIf
	 \EndFor
	 \State \Return unates $\cup$ {\univar}, $\Psi$, {\dependson}

	\end{algorithmic}
\end{algorithm}

%% file: algorithms/cluster.tex
\begin{algorithm}[t]

	\caption{\label{algo:multiclass-clustering}{\clustery}(F(X,Y),k,s,U)}
	\begin{algorithmic}[1]
		\State graph = $\emptyset$\;
		\For {each \emph{clause} of $F(X,Y)$}
			\If {$\langle y_i, y_j \rangle$ pair in \emph{clause}}
				\If {$y_i \not \in U$ and $y_j \not \in U$} 
					\State {\addedge}(graph,$y_i$,$y_j$) \label{line:algo:cluster:addedge}
				\EndIf
			\EndIf
		\EndFor
		\State {\subsety} = $\emptyset$
		\For{$y_i \in Y$ }
			\While { $k \geq 0$}
				\State {\chunk} $\gets$ {\khopneighbor}(graph,$y_i$,k) \label{line:algo:cluster:khop}
				\If {size(\chunk) $\leq s$} \label{line:algo:cluster:tlarge}
					\State break
				\EndIf
				\State $k \gets k-1$ \label{line:algo:cluster:reducek}
			\EndWhile
			\State {\subsety} $\gets$ {\subsety}.add({\chunk}) \label{line:algo:cluster:addchunk}
			\For {$y_j \in$ {\chunk}}
				\State  {\removenode}(graph,$y_j$)\label{line:algo:cluster:removenode}
			\EndFor
		\EndFor
		\State \Return {\subsety}
		
	\end{algorithmic}
\end{algorithm}

%% file: algorithms/classification.tex
\begin{algorithm}
	
	\caption{\label{algo:multiclass-learning}\candidateskf($\Sigma$,F(X,Y),$\Psi$,{\chunk},{\dependson})}
	\begin{algorithmic}[1]
		\State featset $\gets$ X \label{line:algo:classify:featX}
		
		\State D $\gets$ $\emptyset$
		
		\For {each $y_j \in$ Y}
			\For {each $y_i \in$ {\chunk}}
				\If {$y_i \in {\dependson[y_j]}$}
					\State D $\gets$ D $\cup \; y_j$ \label{line:algo:classify:addD}
				\EndIf
			\EndFor
		\EndFor
		
		\For {each $y_j \in$ Y$\setminus$\chunk} 
			\If {$y_j \not \in$ D}
				\State featset $\gets$ featset $\cup$ $y_j$  \label{line:algo:classify:featY}
			\EndIf
		\EndFor
		
		\State feat, lbl $\gets \Sigma_{\downarrow featset}, \; \Sigma_{\downarrow\chunk}$ 
		
		\State dt $\gets$ {\createdt}(feat,lbl)	  \label{line:algo:classify:createtree}
		
		\For {each $y_i \in$ {\chunk}}
			\For {each l $\in$ {\leafnodes}(dt)} \label{line:algo:classify:passtree-start}
					\If {\Label($y_i$,l) =1} \label{line:algo:classify:passtree-label}
						\State $\pi \gets$ Path(dt,root,l) \label{line:algo:classify:passtree-path}
						\State $\psi_i  \gets \psi_i \lor \pi$
					\EndIf
			\EndFor \label{line:algo:classify:passtree-end}
			
			\For {each $y_j \in \psi_i$} \label{line:algo:classify:dependency}
				\State {\dependson[$y_i$]} $\gets$ {\dependson[$y_i$]} $\cup y_j$
			\EndFor \label{line:algo:classify:dependency-end}
		\EndFor
		\State \Return $\Psi$, {\dependson}

	\end{algorithmic}
\end{algorithm}

%% file: algorithms/maxsat.tex
\begin{algorithm}
	
	\caption{\label{algo:maxsat}{\findrepaircandidate}(F(X,Y),$\sigma$,{\totalorder})}
	\begin{algorithmic}[1]
	\State $H \gets F(X,Y) \land (X \leftrightarrow \sigma[X])$ \label{line:algo:maxsat:hard}
	\State $(S,W) \gets \emptyset$
	\For {$y_i \in Y$}\label{line:algo:maxsat:softstart}
		\State $(S,O) \gets (S,O) \cup ((y_i \leftrightarrow \sigma[y'_i]), index({\totalorder}(y_i)))$
	\EndFor \label{line:algo:maxsat:softend}
	\State ind $\gets$ {\maxsatlist}(H,(S,W)) \label{line:algo:maxsat:call}
	\State \Return ind
	\end{algorithmic}
\end{algorithm}

%% file: chapters/example.tex
\subsection{Example}
We now illustrate our algorithm through an example.

\begin{example}
	\label{example:main}
	Let $X=\{x_1,x_2\}$, $Y=\{y_1,y_2,y_3,y_4\}$ in $\exists{Y} F(X,Y)$ where $F(X,Y)$ is $(x_1 \lor x_2 \lor y_1) \land (x_2 \lor \lnot y_1 \lor y_2) \land (y_3 \lor y_4) \land (\lnot y_3 \lor \lnot y_4)$.
\end{example}
\begin{enumerate}
\item%
  {\uniquecall} finds that $y_4$ is defined by $\{x_1,x_2,y_1,y_2, y_3\}$ and returns the Skolem function $\psi_4 = \lnot y_3$. We get $Z = \{y_4\}$ as a determined set.
  
	\item%
	{\tool} generates training data through sampling (Figure \ref{tab:samples_F}). {\tool} attempts to cluster $Y \setminus Z = \{y_1,y_2,y_3\}$ into different chunks of variables to learn together. As $y_1$ and  $y_2$ share a clause, {\clustery} returns the clusters $\{\{y_1,y_2\},\{y_3\}\}$. {\tool} now attempts to learn candidate Skolem functions $\psi_1$, $\psi_2$ together by creating a decision tree (Figure \ref{decisiontree1F}). The decision tree construction uses the samples of $\{x_1,x_2,y_3\}$ as features and samples of $\{y_1,y_2\}$ as labels. The candidate function $\psi_1$ is constructed by taking a disjunction over all paths that end in leaf nodes with label $1$ at index $1$ in the learned decision tree: as shown in Figure \ref{decisiontree1F}, $\psi_1$ is synthesised as $(x_1 \lor (\lnot x_1 \land  \lnot x_2))$. Similarly, considering paths to leaf nodes with label $1$ at index $2$, we get $\psi_2 = ( \lnot x_1 \land \lnot x_2) \lor (\lnot x_1 \land x_2)$, which simplifies to $\lnot x_1$. Now, samples of $\{x_1,x_2,y_1,y_2\}$ are used to predict $y_3$. Considering the path to the leaf node of the learned decision tree with label 1, we get $\psi_3 = x_2$.

	At the end of {\candidateskf}, we have $\psi_1 := (x_1 \lor (\lnot x_1 \land  \lnot x_2)), \; \psi_2 := \lnot x_1, \; \psi_3 := x_2 \;$, and $\psi_4: = \lnot y_3$. Let us assume the total order returned by {\findorder} is $ {\totalorder} = \{y_4, y_3 , y_2 , y_1 \}$.
	\item%
	We construct the error formula, $E(X,Y,Y') = F(X,Y) \land \lnot F(X,Y') \land (Y' \leftrightarrow \Psi)$, which turns out to be {\sat} with counterexample $\sigma = \langle x_1 \leftrightarrow 1$, $x_2 \leftrightarrow 0$, $y_1 \leftrightarrow 0$, $y_2 \leftrightarrow 1$, $y_3 \leftrightarrow 0$, $y_4 \leftrightarrow 1$, $y'_1 \leftrightarrow 1$, $y'_2 \leftrightarrow 0$, $y'_3 \leftrightarrow 0$, $ y'_4 \leftrightarrow 1 \rangle$. 
	
	{\findrepaircandidate} calls {\lexmaxsat} with  $F(X,Y) \land (x_1 \leftrightarrow \sigma[x_1]) \land (x_2 \leftrightarrow \sigma[x_2])$  as hard constraints and $((y_1 \leftrightarrow \sigma[y'_1]),4) \land ((y_2 \leftrightarrow \sigma[y'_2]),3) \land ((y_3 \leftrightarrow \sigma[y'_3]),2) \land ((y_4 \leftrightarrow \sigma[y'_4]),1)$ as soft constraints, with the preference order of soft constraints indicated by their weights. {\findrepaircandidate} returns $ind =\{y_2\}$. 
     Repair synthesis commences for $\psi_2$ with a satisfiability check of $G_2= F(X,Y) \land (x_1 \leftrightarrow \sigma[x_1]) \land (x_2 \leftrightarrow \sigma[x_2]) \land (y_1 \leftrightarrow \sigma[y'_1]) \land (y_2 \leftrightarrow \sigma[y'_2])$. The formula is unsatisfiable, and {\tool} calls {\findcore}, which returns variable $y_1$, since the constraints $(y_1 \leftrightarrow \sigma[y'_1])$ and  $(y_2 \leftrightarrow \sigma[y'_2])$ are not jointly satisfiable in $G_2$.
     As the output $\psi_2$ for the assignment $\sigma$ must change from 0 to 1, $\psi_2$ is repaired by disjoining with $y_1$, and we get $\psi_2:= \lnot x_1 \lor y_1$ as the new candidate.
For the updated candidate vector $\Psi$ the error formula is {\unsat}, and thus $\Psi$ is returned as a Skolem function vector.
\begin{minipage}[10cm]{\textwidth}
	\begin{minipage}[b]{0.26\textwidth}
		\centering
		\begin{tabular}{ccccc}\toprule
			$x_1$ & $x_2$ & $y_1$ & $y_2$ & $y_3$  \\ \midrule
			0 & 0 & 1 & 1 & 0 \\
			0 & 1 & 0 & 1 & 1 \\
			1 & 1 & 1 & 0 & 1 \\ \bottomrule
			
		\end{tabular}
		\captionof{figure}{\label{tab:samples_F}Samples of $F$}
	\end{minipage}
	\hfill
	\begin{minipage}[b]{0.35\textwidth}

		\centering
		\tikz{
			\node[obs,fill=green!10,minimum size=0.7cm] (x1) {$x_1$};%
			\node[latent,below=of x1, minimum size=0.7cm, yshift=0.9cm, xshift=-1cm,fill=green!10] (x2) {$x_2$}; %
			\node[latent, rectangle,  minimum width=0.2cm,
			minimum height = 0.5cm, below=of x1, yshift=0.9cm, xshift= 1cm,fill=red!10] (l1) {$10$}; %
			\node[latent, rectangle, minimum width=0.2cm,
			minimum height = 0.5cm, below=of x2, yshift=0.9cm, xshift= -1cm,fill=red!10] (l2) {$11$}; %
			\node[latent, rectangle,  minimum width=0.2cm,
			minimum height = 0.5cm, below=of x2, yshift=0.9cm, xshift= 1cm,fill=red!10] (l3) {$01$}; %
			\edge {x1}   {l1}
			\edge {x2}   {l2}
			\edge {x2}   {l3}
			
			\path[->,draw]
			(x1) edge node [above] {$0$} (x2)
			(x1) edge node[above] {$1$} (l1)
			(x2) edge node[above] {$0$} (l2)
			(x2) edge node[above] {$1$} (l3)
			
		}
		
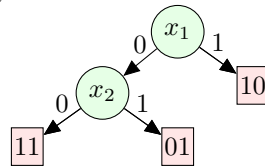
\captionof{figure}{\label{decisiontree1F}Learned decision tree with labels $\{y_1,y_2\}$ and features $\{x_1,x_2,y_3\}$}
		
	\end{minipage}
	\hfill
	\begin{minipage}[b]{0.35\textwidth}

		\centering
		\tikz{
			\node[obs,fill=green!10,minimum size=0.7cm] (x1) {$x_2$};%
			\node[latent, rectangle, minimum width=0.2cm,
			minimum height = 0.5cm, below=of x1, yshift=0.9cm, xshift= 1cm,fill=red!10] (l1) {$1$}; %
			\node[latent, rectangle, minimum width=0.2cm,
			minimum height = 0.5cm, below=of x1, yshift=0.9cm, xshift= -1cm,fill=red!10] (l2) {$0$}; %
			\edge {x1}   {l1}
			\edge {x1}   {l2}
			
			\path[->,draw]
			
			(x1) edge node[above] {$1$} (l1)
			(x1) edge node[above] {$0$} (l2)
			
		}
		
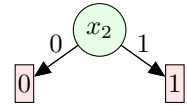
\captionof{figure}{\label{decisiontree2F}Learned decision tree with label $\{y_3\}$ and features $\{x_1,x_2,y_1,y_2\}$}
		
	\end{minipage}
\end{minipage}
\end{enumerate}

%% file: chapters/experimental_results.tex
\section{Experimental Evaluation}
\label{sec:experimental-evaluations}

We conducted an extensive study on $609$ benchmarks that have been previously employed in studies~\cite{AACKRS19,ACGKS18,GRM20}; in particular, we use instances from the 2QBF tracks of QBFEval'17~\cite{qbfeval17} and QBFEval'18~\cite{qbfeval18}, and benchmarks related to arithmetic~\cite{TV17}, disjunctive decomposition~\cite{ACJS17}, and factorization~\cite{ACJS17}. We used Open-WBO~\cite{MML14} for unweighted {\maxsat} queries, RC2~\cite{IMM18} for {\lexmaxsat} queries, and PicoSAT~\cite{B08} to compute {\unsatcore}. Further, we used CryptoMiniSat~\cite{cryptominisat} to find unates and a library based on {\unique}~\cite{F20} to extract unique Skolem functions. We used CMSGen~\cite{GSCM21} to sample the satisfying assignments of the specification.
Finally, we used Scikit-Learn~\cite{sklearn} to learn decision trees and ABC~\cite{abc} to manipulate Boolean functions. All our experiments were conducted on a high-performance computer cluster with each node consisting of a E5-2690 v3 CPU with 24 cores and 96GB of RAM, with a memory limit set to 4GB per core. All tools were run in single-threaded mode on a single core with a timeout of 7200 seconds. We used the \emph{PAR-2} score to compare different techniques, which corresponds to the Penalized Average Runtime, where for every unsolved instance there is a penalty of 2$\times$ timeout.

The objective of our experimental evaluation was to compare the performance of {\tool} with the state-of-the-art tools C2Syn~\cite{AACKRS19}, BFSS~\cite{ACGKS18}, CADET~\cite{R19}, and {\manthan}~\cite{GRM20}, and to analysis the impact of each of the algorithmic modifications implemented in {\tool}. In particular, our empirical evaluation sought answers to the following questions:

\begin{enumerate}
	\item   How does the performance of {\tool} compare with state-of-the-art Skolem functional synthesis tools?
	\item What is the impact on the performance of {\tool} of each of the proposed modifications? %
\end{enumerate}

\paragraph{Summary of Results} {\tool} outperforms all the state-of-the-art tools by solving $509$ benchmarks, while the closest contender, {\manthan}~\cite{GRM20} solves $356$ benchmarks---an increase of $\boldsymbol{153}$ benchmarks over the state-of-the-art. It is worth emphasizing that the increment of 153 is more than twice the improvement shown by {\manthan} over {\cadet}~\cite{R19}, which could solve $280$ benchmarks. %

Moreover, we found that extracting unique functions is useful. There are $246$ benchmarks out of $609$ for which the ratio of $Y$ variables being uniquely defined to the total number of $Y$ is greater than $95 \%$, that is, {\tool} could extract Skolem functions for that many variables via unique function extraction. There is an increase of $25$ benchmarks in the number of solved instances by retaining variables in the determined set to learn and repair candidates. Further, learning candidate functions for a subset of variables together with the help of multi-classification reduces the PAR-2 score from $3227.11$ to $2974.91$. Finally, we see a reduction of $100$ seconds in the PAR-2 score by {\lexmaxsat}.

\subsection{{\tool} vis-a-vis State-of-the-Art Synthesis Tools}\label{sec:experiment-alltool}

\begin{figure}[t]
	\centering
	\includegraphics[scale=0.43]{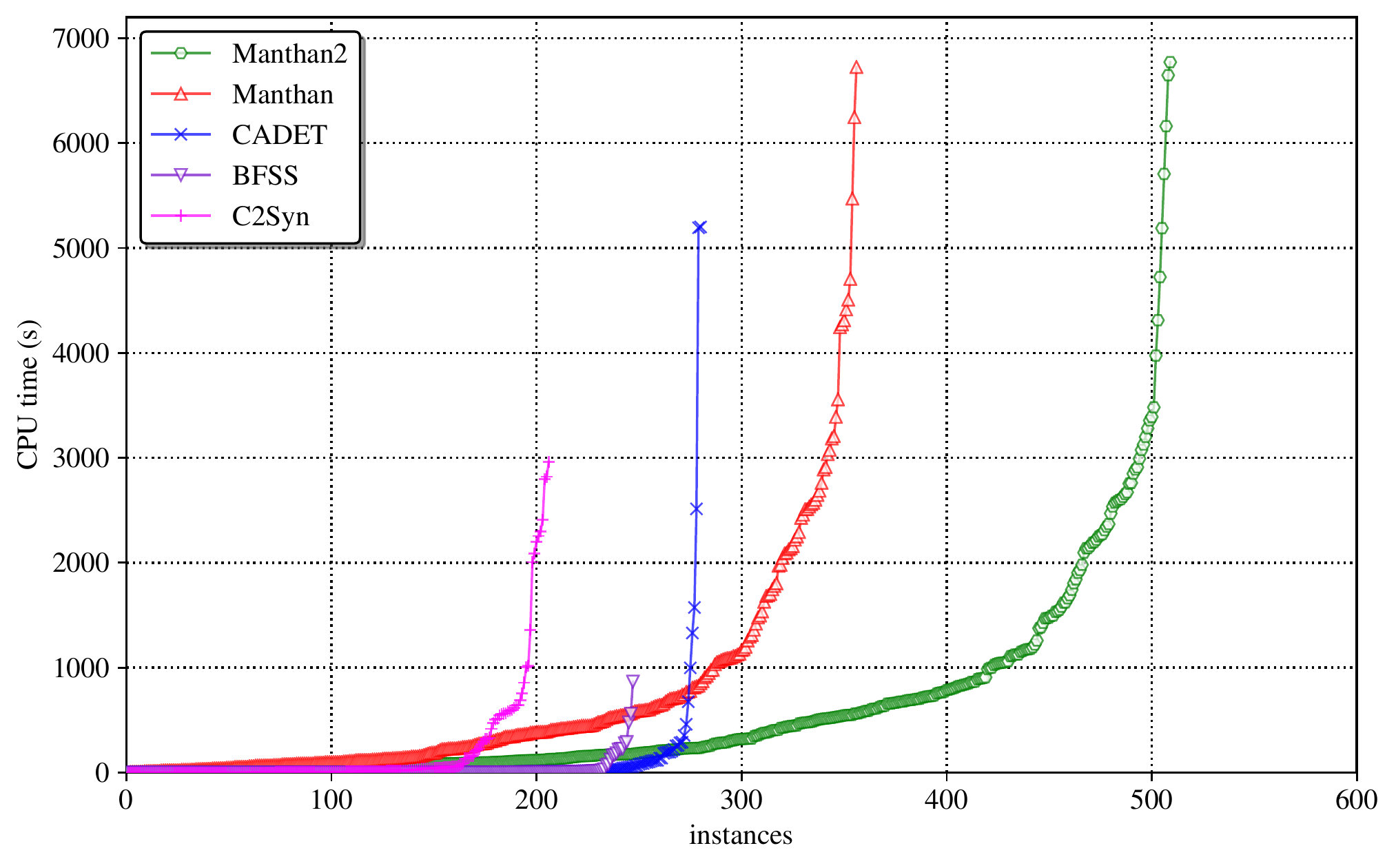}
	\caption{\label{fig:main} \footnotesize Cactus plot: {\tool} vis-a-vis state-of-the-art synthesis tools. Timeout 7200s. Total benchmarks: 609.}
\end{figure}

We compared {\tool} with state-of-the-art tools: {\ctosyn}~\cite{AACKRS19}, {\bfss}~\cite{ACGKS18}, {\cadet}~\cite{R19} and {\manthan}~\cite{GRM20}.  %
Figure~\ref{fig:main} shows a cactus plot to compare the run-time performance of different synthesis tools. The $x$-axis represents the number of benchmarks and $y$-axis represents the time taken, a point $\langle x,y \rangle$ implies that a tool took less than or equal to $y$ seconds to find a Skolem function vector for $x$ many benchmarks out of total $609$ benchmarks. 
\begin{table}[t]
	\centering
	\caption{\label{tab:solved-benchmarks}Performance Summary over 609 benchmarks}
	\begin{tabular}{cccccc}\\ \toprule
		& {\ctosyn} & {\bfss} & {\cadet} & {\manthan} & {\tool} \\ \midrule
		Solved &   206 & 247 & 280 & 356 & 509 \\
		PAR-2  &  9594.83 & 8566.87 & 7817.58 & 6374.39 & 2858.61\\ \bottomrule
	\end{tabular}
\end{table}

\begin{table}[t]
	\centering
	\caption[]{\label{tab:scalmanthan_comparison}{\tool} vs. other state-of-the-art tools. \emph{All tools} represents the union of all state-of-the-art tools.}
	\begin{tabular}{ccccccc}%
		\toprule
		
		\multicolumn{2}{r} {} & {\ctosyn} & {\bfss} & {\cadet} & {\manthan} & All \\ \midrule
		\multirow{2}{*}{{\tool}} & Less & $17$ & $18$ & $21$ & $24$ & $40$ \\ 
		& More & $320$ & $280$ & $250$ & $177$ & $71$ \\ \bottomrule
		
	\end{tabular}
	
\end{table}
As shown in Figure~\ref{fig:main}, {\tool} significantly improves on the state of the art techniques, both in terms of the number of instances solved and runtime performance. In particular, {\tool} is able to solve 509 instances while {\manthan} can solve only 356 instances, thereby achieving an improvement of 153 instances in the number of instances solved. To measure the runtime performance in more detail, we computed PAR-2 scores for all the techniques. The PAR-2 scores for {\tool} and {\manthan} are $2858.61$ and $6374.39$, which is an improvement of $3521.78$ seconds. Finally, we sought to understand if {\tool} performs better than the union of all the other tools. Here, we observe that {\tool} solves $71$ instances that the other tools could not solve, whereas there are only $40$ instances not solved by {\tool} that were solved by one of the other tools.

\subsubsection*{\textbf{{\tool} vis-a-vis {\manthan}:}} 
Table~\ref{tab:manthan2.0-vs-manthan} presents a pairwise comparison of {\tool} with {\manthan}. The first column (PreRepair) presents the number of benchmarks that needed no repair iteration to synthesise a Skolem function vector. The second column (Repair) represents the number of benchmarks that underwent repair iterations. The third column (Self-Sub) presents the number of benchmarks for which at least one variable underwent self-substitution.

We investigate the reason for the increase in the number of benchmarks solved in PreRepair, and observed that {\tool} could extract Skolem functions via unique function extraction for $90\%$ of the variables for $274$ out of these $385$ benchmarks. 

We also observed a significant decrease in the number of benchmarks that needed repair iterations. Out of 124 benchmarks that underwent repair to synthesise a Skolem function vector, only $33$ benchmarks needed self-substitution with {\tool}, whereas there are $75$ out of $224$ benchmarks that needed self-substitution with {\manthan}. The fact that fewer benchmarks required self-substitution to synthesise a Skolem function vector shows that {\tool} could find some hard-to-learn Skolem functions.
\begin{table}[t]
	\centering
	\caption{\label{tab:manthan2.0-vs-manthan} \footnotesize Pairwise comparison of {\tool} with {\manthan}. The table represents the number of benchmarks solved with PreRepair, Repair, and Self-Substitution for {\manthan} and {\tool}.} 
	\begin{tabular}{cccc}
		\\ \toprule
		&  PreRepair & {Repair} & {Self-Sub} \\ \midrule
		{\manthan} & 132 & 224 & 75 \\ 
		{\tool} & 385 & 124 & 33 \\ \bottomrule
		
	\end{tabular}
\end{table}

\subsection{Performance Gain with Each Technical Contribution}
\subsubsection{Impact of Unique Function Extraction}
\begin{figure}[!b]
	\centering
	\includegraphics[scale=0.45]{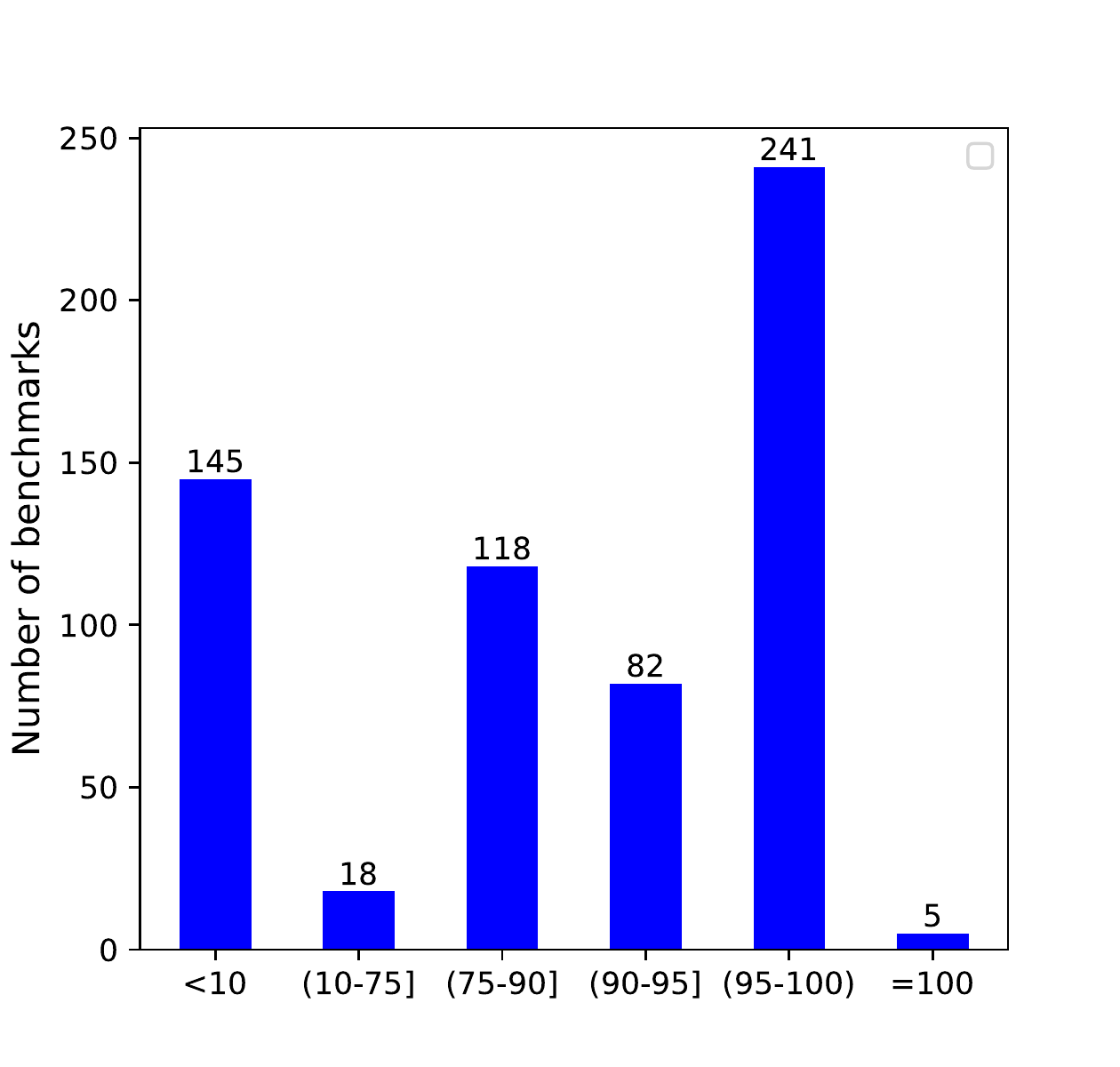}
	\caption{\label{fig:uniqueration} Number of benchmarks by $\%$ ratio of uniquely defined output variables for all 609 benchmarks.}
\end{figure}

We now present the impact of extracting Skolem function for uniquely defined variables. Figure~\ref{fig:uniqueration} shows the percentage of uniquely determined functions on the $x$-axis, and number of benchmarks on $y$-axis. A bar at $x$ shows that $y$ many benchmarks had $x \%$ of $Y$ variables that are uniquely defined. As shown in Figure~\ref{fig:uniqueration}, there are $246$ benchmarks out of $609$ with more than $95\%$ uniquely defined variables; therefore, {\tool} could extract Skolem functions corresponding to these variables via unique function extraction. There are only 5 benchmarks where all the $Y$ variables are defined. Our analysis shows that extracting unique functions significantly reduces the number of $Y$ variables that needed to be learned and repaired in the subsequent phases of {\tool}.

We also analyzed the performance of {\tool} with respect to unique function size. Note that we measure size in terms of number of clauses, as the extracted functions are in CNF. A benchmark is considered to have size $S$ if the maximum size among all its unique functions is $S$. 

Table~\ref{tab:interpolantsize} shows the number of benchmarks with different maximum unique function sizes. There are $136$ benchmarks for which at least one uniquely defined variable has function size greater than 1000 clauses. In general, larger size functions require more data to learn. Table~\ref{tab:interpolantsize} shows that {\tool} was able to extract some hard-to-learn Skolem functions. 

\begin{table}[t]
	\centering
	\caption{\label{tab:interpolantsize} \footnotesize Number of benchmarks with different maximum function size for uniquely defined variables. Function size is measured in terms of number of clauses.} 
	\begin{tabular}{ccccc} \toprule
		& [1-10] & (10-100] & (100-1000] & ($> 1000 $) \\ \midrule
		$\#$-benchmarks & 209  & 203 & 61 & 136 \\ \bottomrule	
	\end{tabular}
\end{table}

An interesting observation is that there were $54$ benchmarks that required self-substitution for just one variable with {\manthan}. However, {\tool} was able to identify that particular variable as uniquely defined and the corresponding function size was more than $3000$ clauses. This observation emphasizes that it is important to extract the functions for uniquely defined variables with large function size in order to efficiently synthesise a Skolem function vector. Therefore, even if there is only one variable with large function size, it is important to extract the corresponding function---the reason for considering maximum size instead of mean or median size in Table~\ref{tab:interpolantsize}.

\subsubsection{Impact of Learning and Repairing over Determined Features}

We now present the impact of variable retention. {\tool} could solved $502$ instances with a PAR-2 score of $3227.11$ by retaining variables in the \emph{determined set} to use them further as features in learning and repairing the other candidates, whereas, if we eliminate them, it could solve only $477$ instances with a PAR-2 score of $3523.28$---a difference of $25$ benchmarks.

It is worth mentioning that there are 370 instances that needed no repair iterations (solved in PreRepair) to synthesise a Skolem function vector when learned with determined features, whereas, if {\tool} does not consider determined features, we see a reduction of 6 benchmark in the number of instances solved in PreRepair.

Interestingly, even if we have fewer such determined features, it is essential to use them to learn and repair the candidates. For example, considering the benchmark \emph{query64\_01}, there are only five variables out of $597$ total $Y$ variables that could be identified as determined features. If we eliminate those five variables, {\tool} could not synthesise a Skolem function vector even with more than $150$ repair iterations within a timeout of 7200s. However, if we retain them as determined features, {\tool} could synthesise a Skolem function vector within 9 repair iterations in less than 400s.

\subsubsection{Efficacy of Multi-Classification and Impact of {\lexmaxsat}}

As discussed in Section~\ref{sec:overview}, two essential questions arise when using multi-classification to learn candidates for a subset of $Y$ together: 1) how to divide the $Y$ variables into different subsets, and 2) how many variables should be learned together?

We experimented with following techniques to divide $Y$ variables into  subsets of sizes $5$ and $8$, i.e, s = 5 or 8:
\begin{enumerate}
	\item Randomly dividing $Y$ variables into different disjoint subsets.
	\item Clustering $Y$ variables in accordance to the edge distance (parameter k) in the primal graph: (i) using $k = 2$ (ii) using $k = 3$
\end{enumerate}

\begin{figure}[!h]
	\centering
	\begin{subfigure}[b]{2.35in}
		\centering
		\includegraphics[scale=0.40]{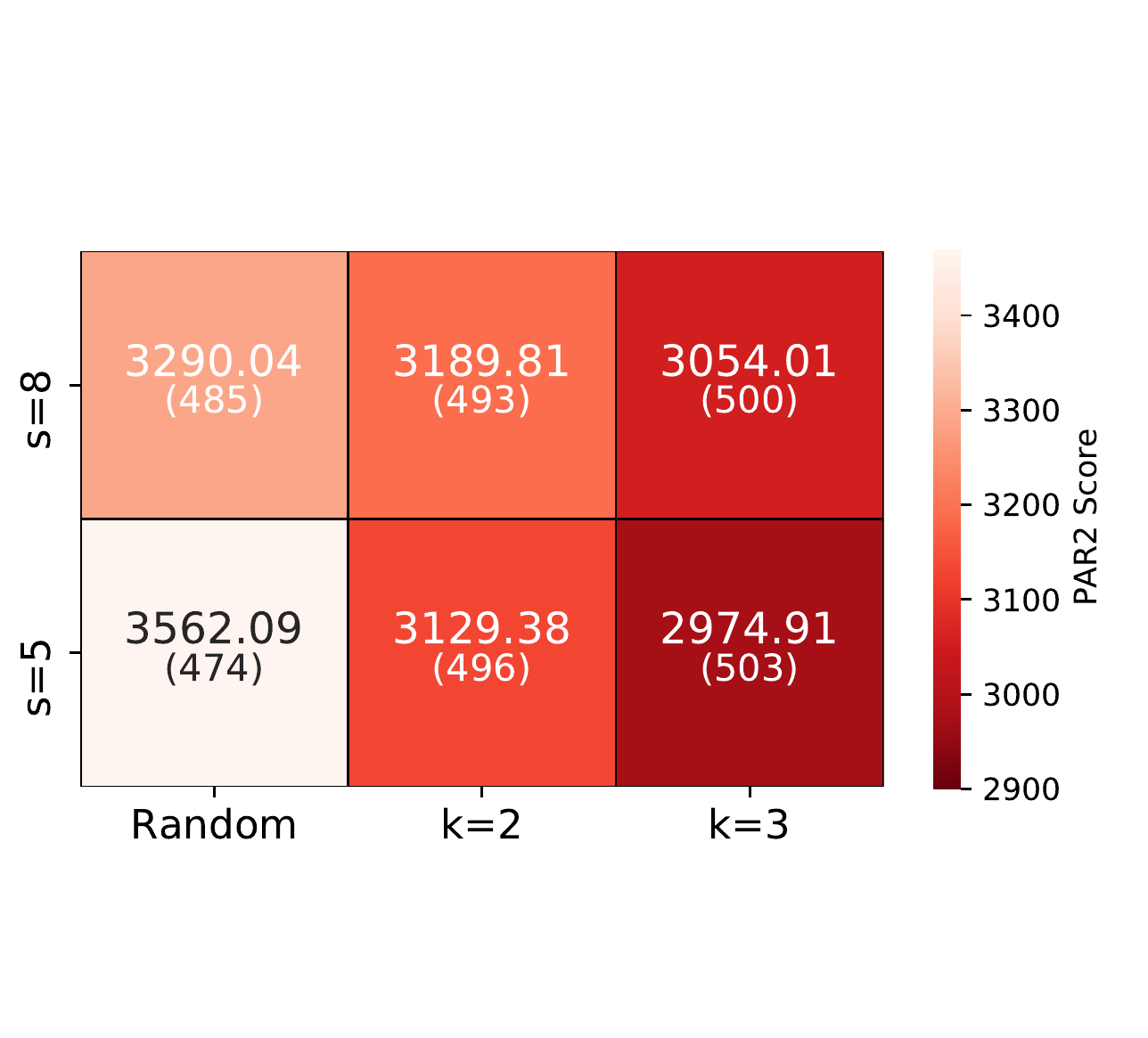}
		\caption{\label{fig:heatmap-multi}{\lexmaxsat} turned off.}		
	\end{subfigure}
	\hfill
	\begin{subfigure}[b]{2.35in}
		\centering
		\includegraphics[scale=0.40]{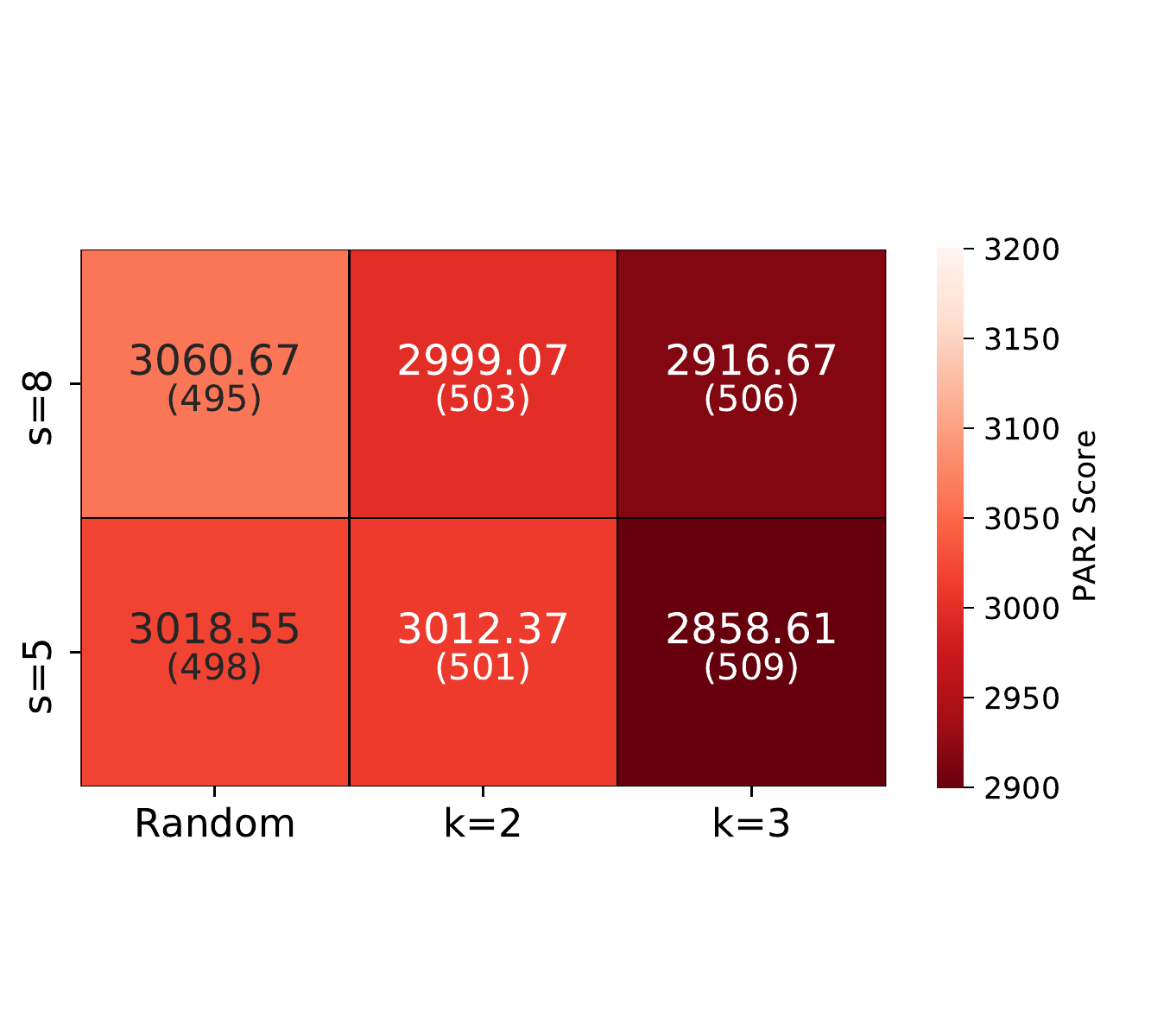}
		\caption{\label{fig:heatmap-lexo} {\lexmaxsat} turned on.}
	\end{subfigure}
	\caption{\label{fig:heatmap} \footnotesize Heatmap of PAR-2 scores achieved by different configuration of {\tool} (darker is better). Here, \emph{s} represents the \emph{size} of sets of variables that were learned together, and $k=2, k=3$ represents the edge distance in the primal graph used to cluster output variables. The number of instances solved by each configuration is shown in brackets. [Best viewed in color].}
\end{figure}

Figure~\ref{fig:heatmap} shows a heatmap of PAR-2 scores for different configurations of {\tool}. A lower PAR-2 score, i.e., a tilt towards the red end of the spectrum in Figure~\ref{fig:heatmap}, indicates a favorable configuration. %
The columns of Figure~\ref{fig:heatmap} correspond to different ways of dividing $Y$ variables into different subsets: (i) \emph{Random}, (ii) $k=2$, and (iii) $k=3$. The rows of Figure~\ref{fig:heatmap} show results for different maximum sizes of such subsets, i.e., s = 5, 8. The number of instances solved in each configuration is also shown in brackets. For comparison, the PAR-2 score of {\tool} with binary classification is 3227.11s and it solved 502 benchmarks.

Let us first discuss Figure~\ref{fig:heatmap-multi}, i.e, the results without {\lexmaxsat}. {\tool} shows a performance improvement with the proposed clustering-based approach in comparison to randomly dividing $Y$ variables into subsets. As shown in Figure~\ref{fig:heatmap-multi}, we observed a drop in PAR-2 score when moving from random to cluster-based partitioning of $Y$ variables. 

We see a better PAR-2 score with graph-based multi-classification compared to binary classification, though the number of instances solved (except with k=3, s=5) is lower than the number of instances solved with binary classification. This shows that dividing $Y$ variables using a cluster-based approach is effective in reducing the candidate learning time. {\tool} performs best with $k=3$ and $s=5$, where it could solve $503$ benchmarks (1 more instance than with binary classification) with a PAR-2 score of $2974.9s$, which amounts to a reduction of $252$ seconds over the PAR-2 score with binary classification. We observe a similar trend with {\lexmaxsat} turned on (as shown in Figure~\ref{fig:heatmap-lexo}).

Finally, let us move our attention towards the impact of {\lexmaxsat}, shown in Figure~\ref{fig:heatmap-lexo}. {\tool} uses {\lexmaxsat} only if the number of candidates to repair exceeds 50 times the number of candidates chosen by {\maxsat}. A comparison of  Figure~\ref{fig:heatmap-multi} and Figure~\ref{fig:heatmap-lexo} shows that with {\lexmaxsat}, {\tool} solves at least 3 more benchmarks for all the configurations.

{\tool} performs best when we turn on {\lexmaxsat} and set $k=3$ as well as $s=5$. The results discussed in Section~\ref{sec:experiment-alltool} were achieved $k=3$ and $s=5$.

%% file: chapters/relatedwork.tex
\section{Related Work}
\label{sec:relatedwork}
Boolean functional synthesis is a classical problem. Its origin traces back to Boole's seminal work~\cite{boole1847}, which was subsequently pursued with a focus on decidability---by L\"owenheim and Skolem~\cite{L1910}.

The past decade has seen significant progress in the development of efficient tools for Boolean functional synthesis, driven by a diverse set of techniques.
Quantifier elimination by functional composition can be an efficient approach when paired with Craig interpolation to reduce the size of composite functions~\cite{JLH09,J09}.
However, interpolation does not reliably find succinct composite functions, thus limiting scalability of this method.
More recently, it was shown that ROBDDs lend themselves well to functional composition~\cite{FTV16} (even without interpolation) and they can take advantage of factored specifications~\cite{TV17}.

Instead of directly deducing Skolem functions from a specification, a series of CEGAR-based synthesis algorithms start from an initial set of approximate functions that are rectified in a subsequent phase of counterexample guided refinement~\cite{JSCTA15,ACJS17,ACGKS18}. It was observed that the initial functions are often valid Skolem functions~\cite{ACGKS18}. This naturally leads to the question as to which classes of specifications admit efficient Boolean functional synthesis, which has recently been studied from the area of knowledge compilation~\cite{ACGKS18,AACKRS19}.

So-called incremental determinization can be seen as lifting Conflict-Driven Clause Learning (CDCL) to the level of Boolean functions~\cite{RS16,RTRS18,R19}: variables with unique Skolem functions are successively identified, in analogy with unit propagation, and whenever this process comes to a halt, a Skolem function for one of the remaining variables is fixed by adding auxiliary clauses.
While originally developed as a decision procedure for 2QBF, the algorithm was later successfully adapted to perform functional synthesis for non-valid specifications~\cite{R19}.

Skolem functions can also be efficiently extracted from proofs generated by QBF solvers~\cite{BJ12,NPLSB12,HSB14,RT15,BJJW15,SSWZ20}, but this requires both a valid input specification and a proof of validity (which itself is typically hard to compute).

Recently, a data-driven approach to Boolean functional synthesis was proposed~\cite{GRM20}.
Data-driven approaches have  proven to be efficient for the other forms of synthesis, like invariant synthesis~\cite{ENDGM18,GLMN14,GLST05}, or synthesis by example~\cite{FG19}.

Our data-driven approach benefits from identifying variables that are defined by a subset of input variables, since the corresponding definitions represent Skolem functions that do not have to be learned.
Such definitions are often introduced as an artifact of converting circuits into CNF formulas, where gates are encoded by auxiliary variables that are defined in term of their inputs.
Standard techniques for recovering gate definitions from CNF formulas (some of which are also used in Boolean synthesis tools~\cite{ACGKS18,AACKRS19}) rely on pattern matching of clauses and variables induced by specific gate types~\cite{RMB04,FM07,GB13}. These methods are fast but can only detect definitions from a pre-defined library of gates. By contrast, {\tool} extracts the functions for uniquely defined variables using semantic gate extraction based on propositional interpolation~\cite{F20}. This approach is computationally more expensive (each definability check requires a SAT call), but it is complete: whenever a variable y is defined in terms of a given set X of variables, the corresponding definition will be returned.

%% file: chapters/conclusion.tex
\section{Conclusion}
\label{sec:conclusion}
Boolean functional synthesis a fundamental problem with many applications. In this paper, we showed how to improve the state-of-the-art data-driven Skolem function synthesiser {\manthan} to achieve better scalability. We proposed crucial algorithm innovation, and used them in a new framework, called {\tool}. In particular, the proposed modifications are: computing unique Skolem functions by definition extraction, retaining variables with Skolem functions as \emph{determined features} instead of eliminating them, using multi-classification to jointly learn candidate functions for sets of output variables, and using {\lexmaxsat} to reduce the number of repair iterations. With these proposed improvements, {\tool} could synthesise a Skolem function vector for $509$ instances out a total of $609$, compared to $356$ instances solved by {\manthan}.

\textbf{Acknowledgments:} This work was supported in part by National Research Foundation Singapore under its NRF Fellowship Programme [NRF-NRFFAI1-2019-0004 ] and AI Singapore Programme [AISG-RP-2018-005], NUS ODPRT Grant [R-252-000-685-13], and the Vienna Science and Technology Fund (WWTF) [ICT19-060]. The computational work for this article was performed on resources of the National Supercomputing Center, Singapore: \url{https://www.nscc.sg}.

%% file: chapters/appendix.tex
\section*{Appendix}
\subsection*{Unique Function Extraction: Additional Experiments}
	\begin{figure}
		\centering
			\includegraphics[scale=0.45]{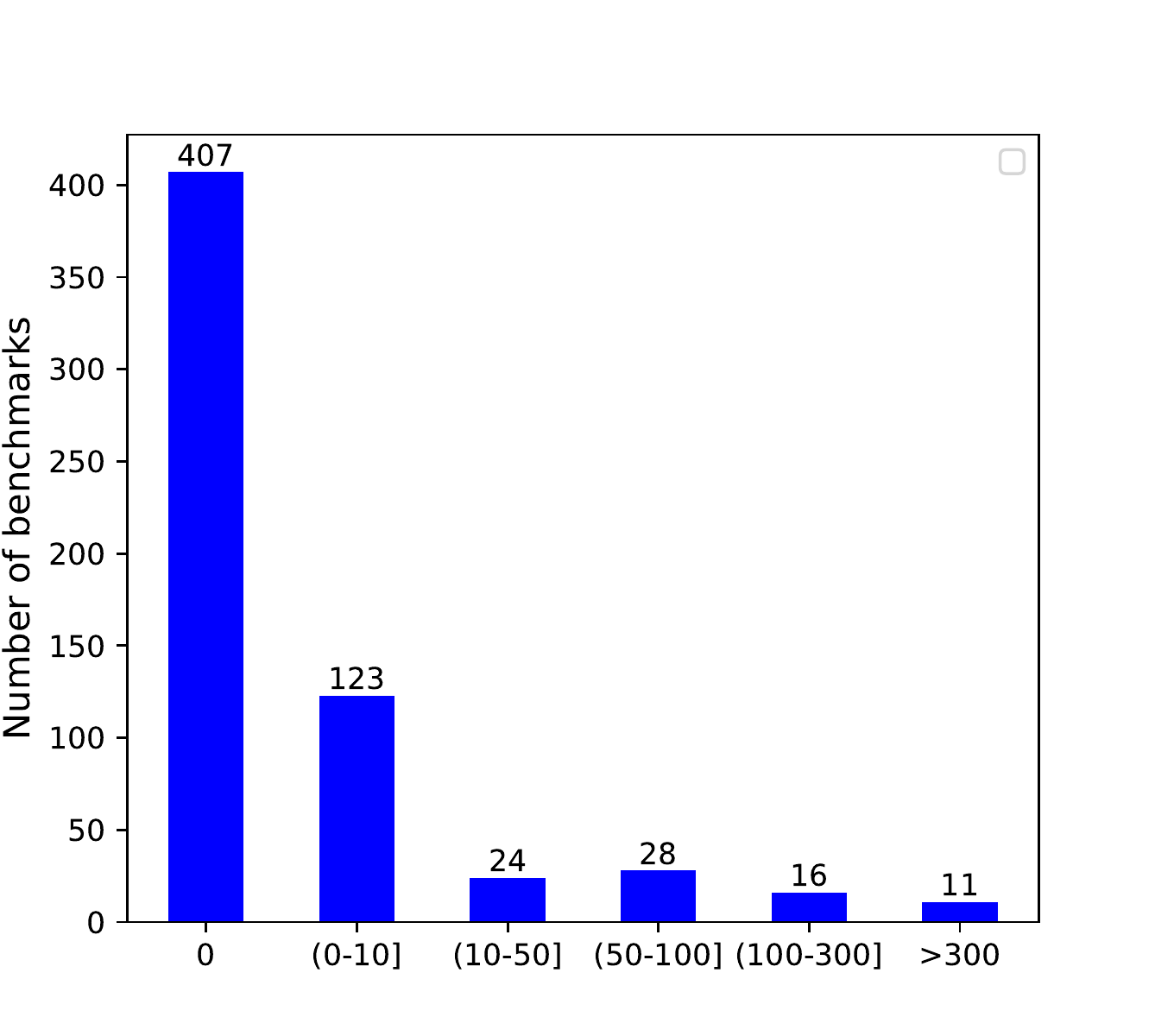}
			\captionof{figure}{\label{fig:unique-over-bfss}Plot to represent number of benchmarks 
				concerning $\%$ gain in number of uniquely defined Y variables with semantic over syntactic gate extraction for all 609 benchmarks.}
		
	\end{figure}
	Figure~\ref{fig:unique-over-bfss} shows the advantage of using interpolation-based extraction (used by {\tool}) versus the simpler syntactic gate extraction technique~~\cite{RMB04,FM07,GB13}. Figure~\ref{fig:unique-over-bfss} shows the $\%$ increment on $x-axis$ and number of benchmarks on $y-axis$. A bar at $x$ shows that for $y$ many benchmarks, semantic gate extraction has found $x\%$ of more unique defined variables than that of syntactic gate extraction.  11 benchmarks have more than $300\%$ $Y$ variables that were identified as uniquely defined variables by interpolation-based extraction over syntactic gate extraction. An interesting observation is, there were $54$ benchmarks that needed self-substitution as fall-back for just one variable with {\manthan}. However, {\tool} was able to identify that particular variable as uniquely defined, but with definition size more than $3000$. Note that, we measure the definition size in terms of number of clauses. All of these $54$ benchmarks falls in the category of $\leq 10\%$ increment with semantic gate extraction over syntactic. This observation proves that it is important to identify the uniquely defined variables with large definition size in order to efficiently synthesise a Skolem function.

	We also did an experiment with limit on function size, that is, the Skolem function for a uniquely defined $y$  variable is extracted only if the function size is greater than $10$ clauses, but less than $1000$ clause. In particular, the objective of this experiment to see if {\tool} can efficiently learn the candidate functions for the variables that are uniquely defined and have either very small or very large function size. 
	
	It turns-out that {\tool} needs to extract functions for all uniquely variables found by interpolation based extraction to perform better irrespective of their function sizes. If {\tool} does not extract the function with size less than $1000$, then out of $264$ benchmarks(column 2 and 3 of Table~\ref{tab:interpolantsize}), {\tool} can solve only $77$ benchmarks. Similarly, if {\tool} does not extract the function with size greater than 1000, then {\tool} can solve only $26$ out of $136$ benchmarks(column 5 of Table~\ref{tab:interpolantsize}). The one possible reasoning for this behavior can be that {\tool} has difficulty in learning \emph{good} candidate functions when there are too many uniquely defined variables with small functions, or there are a few uniquely defined variable with large functions.

%% file: cav21_manthan.bbl
\begin{thebibliography}{10}
\providecommand{\url}[1]{\texttt{#1}}
\providecommand{\urlprefix}{URL }
\providecommand{\doi}[1]{https://doi.org/#1}

\bibitem{qbfeval17}
{QBF} solver evaluation portal 2017, \url{http://www.qbflib.org/qbfeval17.php}

\bibitem{qbfeval18}
{QBF} solver evaluation portal 2018, \url{http://www.qbflib.org/qbfeval18.php}

\bibitem{sklearn}
sklearn.tree.decisiontreeclassifier,
  \url{https://scikit-learn.org/stable/modules/generated/sklearn.tree.DecisionTreeClassifier.html}

\bibitem{AACKRS19}
Akshay, S., Arora, J., Chakraborty, S., Krishna, S., Raghunathan, D., Shah, S.:
  Knowledge compilation for {B}oolean functional synthesis. In: Proc. of FMCAD
  (2019)

\bibitem{ACGKS18}
Akshay, S., Chakraborty, S., Goel, S., Kulal, S., Shah, S.: What’s hard about
  {B}oolean functional synthesis? In: Proc. of CAV (2018)

\bibitem{ACJS17}
Akshay, S., Chakraborty, S., John, A.K., Shah, S.: Towards parallel {B}oolean
  functional synthesis. In: Proc. of TACAS (2017)

\bibitem{ABGL13}
Ans{\'o}tegui, C., Bonet, M.L., Gabas, J., Levy, J.: Improving wpm2 for
  (weighted) partial max{SAT}. In: Proc. of CP (2013)

\bibitem{BJ12}
Balabanov, V., Jiang, J.H.R.: Unified {QBF} certification and its applications.
  In: Proc. of FMCAD (2012)

\bibitem{BJJW15}
Balabanov, V., Jiang, J.R., Janota, M., Widl, M.: Efficient extraction of {QBF}
  (counter)models from long-distance resolution proofs. In: Proc. of {AAAI}
  (2015)

\bibitem{B08}
Biere, A.: {PicoSAT} essentials. Proc. of JSAT  (2008)

\bibitem{BLS11}
Biere, A., Lonsing, F., Seidl, M.: Blocked clause elimination for {QBF}. In:
  Proc. of CADE (2011)

\bibitem{boole1847}
Boole, G.: The mathematical analysis of logic. Philosophical Library (1847)

\bibitem{B89}
Brayton, R.K.: Boolean relations and the incomplete specification of logic
  networks. In: Proc. of VLSID (1989)

\bibitem{BS89}
Brayton, R.K., Somenzi, F.: An exact minimizer for boolean relations. In: Proc.
  of ICCAD (1989)

\bibitem{ENDGM18}
Ezudheen, P., Neider, D., D'Souza, D., Garg, P., Madhusudan, P.: Horn-{ICE}
  learning for synthesizing invariants and contracts. In: Proc. of OOPSLA
  (2018)

\bibitem{FG19}
Fedyukovich, G., Gupta, A.: Functional synthesis with examples. In: Proc. of CP
  (2019)

\bibitem{FTV16}
Fried, D., Tabajara, L.M., Vardi, M.Y.: {BDD}-based {B}oolean functional
  synthesis. In: Proc. of CAV (2016)

\bibitem{FM07}
Fu, Z., Malik, S.: Extracting logic circuit structure from conjunctive normal
  form descriptions. In: Proc. of VLSID (2007)

\bibitem{GLMN14}
Garg, P., L{\"o}ding, C., Madhusudan, P., Neider, D.: {ICE}: A robust framework
  for learning invariants. In: Proc. of CAV (2014)

\bibitem{GDN92}
Ghosh, A., Devadas, S., Newton, A.R.: Heuristic minimization of boolean
  relations using testing techniques. IEEE transactions on computer-aided
  design of integrated circuits and systems  (1992)

\bibitem{GRM20}
Golia, P., Roy, S., Meel, K.S.: Manthan: A data driven approach for {B}oolean
  function synthesis. In: Proc. of CAV (2020)

\bibitem{GSCM21}
Golia, P., Soos, M., Chakraborty, S., Meel, K.S.: Designing samplers is easy:
  The boon of testers. In: Proc. of FMCAD (2021)

\bibitem{GB13}
Goultiaeva, A., Bacchus, F.: Recovering and utilizing partial duality in {QBF}.
  In: Proc. of {SAT} (2013)

\bibitem{GLST05}
Grumberg, O., Lerda, F., Strichman, O., Theobald, M.: Proof-guided
  underapproximation-widening for multi-process systems. In: Proc. of POPL
  (2005)

\bibitem{HSB14}
Heule, M.J., Seidl, M., Biere, A.: Efficient extraction of {S}kolem functions
  from {QRAT} proofs. In: Proc. of FMCAD (2014)

\bibitem{IMM18}
Ignatiev, A., Morgado, A., Marques{-}Silva, J.: {PySAT:} {A} {Python} toolkit
  for prototyping with {SAT} oracles. In: Proc. of SAT (2018)

\bibitem{JS92}
Jeong, S.W., Somenzi, F.: A new algorithm for the binate covering problem and
  its application to the minimization of boolean relations. In: Proc. of ICCAD
  (1992)

\bibitem{J09}
Jiang, J.H.R.: Quantifier elimination via functional composition. In: Proc. of
  CAV (2009)

\bibitem{JLH09}
Jiang, J.R., Lin, H., Hung, W.: Interpolating functions from large boolean
  relations. In: Proc. of {ICCAD}. pp. 779--784. {ACM} (2009)

\bibitem{JMF14}
Jo, S., Matsumoto, T., Fujita, M.: {SAT}-based automatic rectification and
  debugging of combinational circuits with lut insertions. Proc. of IPSJ T-SLDM
   (2014)

\bibitem{JSCTA15}
John, A.K., Shah, S., Chakraborty, S., Trivedi, A., Akshay, S.: Skolem
  functions for factored formulas. In: Proc. of FMCAD (2015)

\bibitem{KS00}
Kukula, J.H., Shiple, T.R.: Building circuits from relations. In: Proc. of CAV
  (2000)

\bibitem{LM08}
Lang, J., Marquis, P.: On propositional definability. Artificial Intelligence
  (2008)

\bibitem{LS90}
Lin, B., Somenzi, F.: Minimization of symbolic relations. In: Proc. of ICCAD
  (1990)

\bibitem{abc}
Logic, B., Group, V.: {ABC}: A system for sequential synthesis and
  verification, \url{http://www.eecs.berkeley.edu/~alanmi/abc/}

\bibitem{L1910}
L\"{o}wenheim, L.: \"{U}ber die {A}ufl{\"o}sung von {G}leichungen im logischen
  {G}ebietekalkul. Mathematische Annalen  (1910)

\bibitem{MAGL11}
Marques-Silva, J., Argelich, J., Gra{\c{c}}a, A., Lynce, I.: {B}oolean
  lexicographic optimization: algorithms \& applications. Proc. of Annals of
  Mathematics and Artificial Intelligence  (2011)

\bibitem{MML14}
Martins, R., Manquinho, V., Lynce, I.: {Open-WBO}: A modular {MaxSAT} solver.
  In: Proc. of SAT (2014)

\bibitem{MM20}
Massacci, F., Marraro, L.: Logical cryptanalysis as a {SAT} problem. Journal of
  Automated Reasoning  (2000)

\bibitem{NPLSB12}
Niemetz, A., Preiner, M., Lonsing, F., Seidl, M., Biere, A.: Resolution-based
  certificate extraction for {QBF}. In: Proc. of SAT (2012)

\bibitem{R19}
Rabe, M.N.: Incremental determinization for quantifier elimination and
  functional synthesis. In: Proc. of CAV (2019)

\bibitem{RS16}
Rabe, M.N., Seshia, S.A.: Incremental determinization. In: Proc. of {SAT}
  (2016)

\bibitem{RT15}
Rabe, M.N., Tentrup, L.: {CAQE}: A certifying {QBF} solver. In: Proc. of FMCAD
  (2015)

\bibitem{RTRS18}
Rabe, M.N., Tentrup, L., Rasmussen, C., Seshia, S.A.: Understanding and
  extending incremental determinization for {2QBF}. In: Proc. of CAV (2018)

\bibitem{RMB04}
Roy, J.A., Markov, I.L., Bertacco, V.: Restoring circuit structure from {SAT}
  instances. In: Proc. of IWLS (2004)

\bibitem{SS10}
Samer, M., Szeider, S.: Algorithms for propositional model counting. Journal of
  Discrete Algorithms  (2010)

\bibitem{SSWZ20}
Schlaipfer, M., Slivovsky, F., Weissenbacher, G., Zuleger, F.: Multi-linear
  strategy extraction for {QBF} expansion proofs via local soundness. In: Proc.
  of {SAT} (2020)

\bibitem{F20}
Slivovsky, F.: Interpolation-based semantic gate extraction and its
  applications to {QBF} preprocessing. In: Proc. of CAV (2020)

\bibitem{cryptominisat}
Soos, M.: msoos/cryptominisat (2019),
  \url{https://github.com/msoos/cryptominisat}

\bibitem{SGM20}
Soos, M., Gocht, S., Meel, K.S.: Tinted, detached, and lazy {CNF-XOR} solving
  and its applications to counting and sampling. In: Proc. of CAV (2020)

\bibitem{SGF13}
Srivastava, S., Gulwani, S., Foster, J.S.: Template-based program verification
  and program synthesis. STTT  (2013)

\bibitem{TV17}
Tabajara, L.M., Vardi, M.Y.: Factored {B}oolean functional synthesis. In: Proc.
  of FMCAD (2017)

\end{thebibliography}
